\PassOptionsToPackage{libertine,vvarbb}{newtxmath}
\documentclass[a4paper]{article}
\usepackage[final]{colm2025_conference}

\setcitestyle{numbers,square,comma,sort,compress}

\usepackage{xspace}
\usepackage{url}
\usepackage{tabularray}
\UseTblrLibrary{booktabs}
\usepackage{amsmath}
\usepackage{algorithm}
\usepackage{algpseudocode}
\usepackage{graphicx}
\usepackage{subcaption}
\usepackage{xcolor}
\usepackage{colortbl}
\usepackage{multirow}
\usepackage{adjustbox}
\usepackage{geometry}
\usepackage{pdflscape}
\usepackage{lineno}
\usepackage{placeins}
\usepackage{setspace}
\usepackage{threeparttable}
\usepackage{hyperref}
\usepackage{cleveref}

\graphicspath{{./fig/}}

\definecolor{openweight}{RGB}{255, 245, 230}  %
\definecolor{fullyopen}{RGB}{245, 235, 255}   %
\definecolor{ours}{RGB}{235, 255, 235}        %

\definecolor{openweight}{RGB}{255, 245, 230} %
\definecolor{fullyopen}{RGB}{245, 235, 255}  %
\definecolor{ours}{RGB}{235, 255, 235}       %

\newcommand{\sys}{\textsc{Kaiyuan-2B}\xspace}
\newcommand{\fullsys}{\textsc{PCMind-2.1-Kaiyuan-2B}\xspace}
\newcommand{\dataframework}{\textsc{Kai\-yuan-Spark}\xspace}
\newcommand{\papertitle}{\fullsys Technical Report}

\let\tsup\textsuperscript

\hypersetup{
  pdftitle={\papertitle},
  pdfauthor={Kairong Luo, Zhenbo Sun, Xinyu Shi, Shengqi Chen, Bowen Yu, Yunyi Chen, Chenyi Dang, Hengtao Tao, Hui Wang, Fangming Liu, Kaifeng Lyu, Wenguang Chen},
}

\fancypagestyle{firstpage}{%
  \fancyhf{}
  \fancyfoot[C]{\thepage}
  \renewcommand{\headrulewidth}{0pt}
}

\crefname{section}{\S}{\S\S}
\Crefname{section}{Section}{Sections}

\NewTblrEnviron{narrowtblr}
\SetTblrInner[narrowtblr]{
  hline{1,Z} = {1pt},
  hline{2}   = {0.5pt},
  rows = {rowsep=0.8pt},
  rowhead = 1,
  rows = {valign=m},
  row{1} = {font=\bfseries, bg=gray!10},
}

\title{\papertitle}

\usepackage[tt=false]{libertine} %
\makeatletter
\let\@maketitleold\@maketitle
\def\@maketitle{%
\fancyheadoffset{0.5cm}%
\@maketitleold%
\lhead{\papertitle}%
\rhead{\leftmark}%
}
\makeatother

\author{
Kairong Luo\tsup{1}\thanks{luokr24@mails.tsinghua.edu.cn, sunzb20@mails.tsinghua.edu.cn}, 
Zhenbo Sun\tsup{1}\footnotemark[1],
Xinyu Shi\tsup{1}, Shengqi Chen\tsup{1}, Bowen Yu\tsup{3}, Yunyi Chen\tsup{1}, Chenyi Dang\tsup{1},\\
\bf\ 
Hengtao Tao\tsup{2},
Hui Wang\tsup{2},
Fangming Liu\tsup{2}, 
Kaifeng Lyu\tsup{1},
Wenguang Chen\tsup{1,2}\setcounter{footnote}{1}\thanks{Corresponding authors.}\\
\tsup{1}Tsinghua University, \tsup{2}Peng Cheng Laboratory, \tsup{3}Beijing Houdu Technology Co., Ltd%
}

\begin{document}

\maketitle
\thispagestyle{firstpage}

\begin{abstract}

The rapid advancement of Large Language Models (LLMs) has resulted in a significant knowledge gap between the open-source community and industry, primarily because the latter relies on closed-source, high-quality data and training recipes. To address this, we introduce \textbf{\fullsys}, a fully open-source 2-billion-parameter model focused on improving training efficiency and effectiveness under resource constraints. Our methodology includes three key innovations: a \emph{Quantile Data Benchmarking} method for systematically comparing heterogeneous open-source datasets and providing insights on data mixing strategies; a \emph{Strategic Selective Repetition} scheme within a multi-phase paradigm to effectively leverage sparse, high-quality data; and a \emph{Multi-Domain Curriculum Training} policy that orders samples by quality. Supported by a highly optimized data preprocessing pipeline and architectural modifications for $\text{FP}16$ stability, \sys achieves performance competitive with state-of-the-art fully open-source models, demonstrating practical and scalable solutions for resource-limited pretraining. We release all assets (including model weights, data, and code) under Apache 2.0 license at \url{https://huggingface.co/thu-pacman/PCMind-2.1-Kaiyuan-2B}.

\end{abstract}

\vspace{-0.2in}
\section{Introduction}

\begin{figure}[!ht]
\vspace{-0.2in}
\centering
\includegraphics[width=0.95\linewidth]{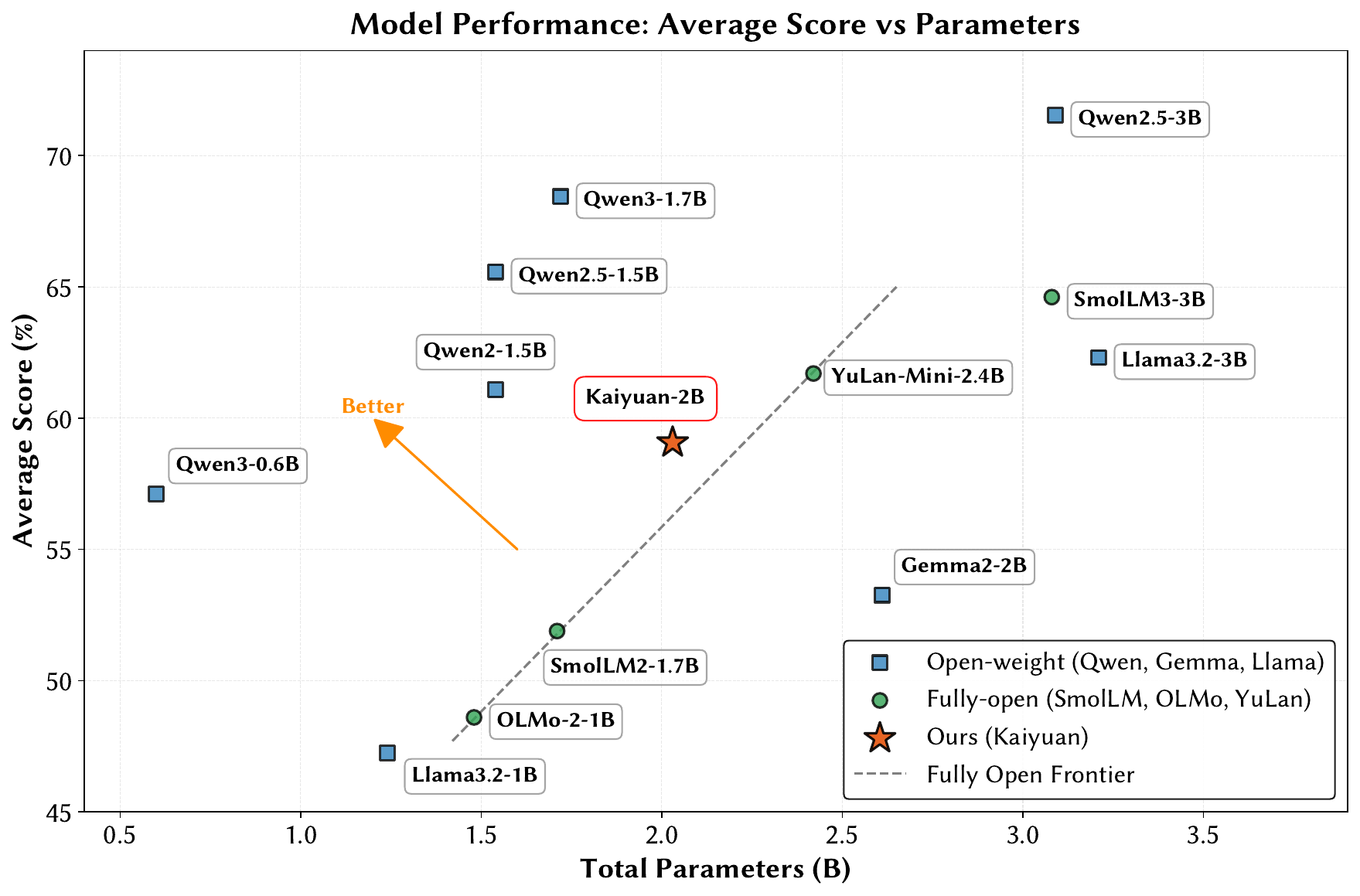}
\vspace{-0.1in}
\caption{Model Performance Comparison. \sys surpasses the frontier of fully open-source models at a similar scale, and approaches open-weight models such as Qwen2-1.5B~\citep{qwen2} and Llama3.2-3B~\citep{llama3.2}. A full version of the corresponding benchmark scores is detailed in \Cref{tab:model_comparison_full}.}
\label{fig:model_perf_comp}
\end{figure}

The field of Large Language Models (LLMs) has seen remarkable advancements, demonstrating comprehensive capabilities across a wide spectrum of tasks. The performance of these models fundamentally depends on the quality and scale of their pretraining~\citep{yue2025does,hernandez2021transfer}. However, the core science and engineering behind large-scale LLM pretraining remain underexplored by the academic and open-source communities due to two main industry practices: (1) \textbf{Closed-source pretraining} of leading models~\citep{openai2023gpt4,gemini2025gemini25}. (2) Release of \textbf{open model weights} but with \textbf{closed-source training recipes}~\citep{deepseekai2025deepseekv3technicalreport,deepseekai2025deepseekr1incentivizingreasoningcapability,qwen3}.

\noindent \textbf{Fully open-source models}, which publish both weights, datasets, and detailed pretraining procedures, are essential to bridge this knowledge gap and facilitate academic exploration. Pioneer works in this direction include the OLMo series~\citep{groeneveld-etal-2024-olmo,olmo20252olmo2furious,muennighoff2025olmoe}, SmolLM series~\citep{smol2,smollm3}, and Yulan series~\citep{zhu2024yulan,hu2024yulan}. Furthermore, the increasing availability of high-quality, open-source pretraining datasets, spanning English, multilingual, code, and math domains~\citep{fineweb,DCLM,stack-v2,kydlicek2025finepdfs,together2023redpajama,fineweb-edu-chinese,zhou2025megamath}, lays a crucial foundation for more open pretraining attempts.

Despite these advancements, significant challenges persist in open-source pretraining, particularly when attempting to match the performance of state-of-the-art open-weight models under resource constraints. In this technical report, we introduce the \textbf{\fullsys (\sys in short), a fully open-source model}, and detail its pretraining methodology. Our work focuses on two critical challenges faced by resource-limited communities:

\begin{enumerate}
\item \textbf{Heterogeneous Open-Source Data:} While many pretraining-scale datasets are available, their sources and preprocessing pipelines vary significantly~\citep{DCLM,fineweb,stack-v2}. This leads to vast differences in data features, posing a challenge for the effective comparison, selection, and mixing of these heterogeneous datasets~\citep{su2025nemotroncctransformingcommoncrawl}.
\item \textbf{Limited Compute Resources:} The academic community typically cannot afford to train on the scale of tokens (e.g., tens of trillions) used by industry leaders~\citep{qwen3}. This necessitates novel strategies to improve training efficiency with limited data and computational resources.
\end{enumerate}

\noindent  Our primary goal is to push the frontier of open-source pretraining by directly addressing these two questions:
\begin{enumerate}
\item How can one properly compare, select, and effectively mix heterogeneous open-source datasets?
\item How to improve the training efficiency, especially when dealing with the inherent sparsity of high-quality data?
\end{enumerate}

In the pretraining of the \sys model, we propose and implement practical solutions to these challenges, centered on data management and training efficiency. Given the domains of accessible open-source datasets, we focus on improving the model's Chinese, coding, and math capabilities, in addition to its general knowledge and reasoning capabilities. As shown in \Cref{fig:model_perf_comp}, through these practices, our \sys model demonstrates competitive performance within the fully open-source category and narrows the gap to open-weight models. Our contributions include the following aspects:

\paragraph{Deduplication and Quantile Data Benchmarking.}
We propose a novel \textbf{quantile benchmarking} method to systematically evaluate and compare leading open-source datasets (e.g., DCLM-Baseline~\citep{DCLM}, Fineweb-Edu~\citep{fineweb}). The rationale is twofold: (1) Open-source datasets often include rule-based or model-based quality metrics, which have proven effective in filtering and can be used to inform the importance of samples during comparison~\citep{fineweb,DCLM}. (2) By selecting a data subset around a target quality score quantile and training a small reference model over it, we can measure the subset's characteristics via the reference model's downstream performance. This method allows us to understand how different datasets or distinct partitions within a single dataset perform across various capabilities. The approach enables systematic benchmarking across heterogeneous collections, especially for the leading datasets that account for the majority of training tokens. Crucially, deduplication is performed before the quantile benchmarking process. (\cref{sec:data_benchmarking_subsec})

\paragraph{System Infrastructure and Training Stability.}
To support these data-centric efforts, we built a high-performance and scalable data preprocessing pipeline based on Spark~\citep{zaharia2012spark} and optimized with the Chukonu framework~\citep{chukonu}. This optimized framework efficiently supports deduplication and leverages Spark's native sorting for curriculum implementation.
Finally, our pretraining experiments were conducted on Ascend 910A clusters. Ascend 910A is comparable to V100 hardware and supports only $\text{FP}16$. To ensure training stability under these conditions, we modify the model architecture based on Qwen3-1.7B by incorporating sandwich normalization and soft capping, in addition to standard $\text{QK}$ normalization. (\cref{sec:architecture_stability,sec:data_preprocessing_framework})

\paragraph{Strategic Selective Repetition for Sparse High-Quality Data.}
Our data benchmarking confirms that high-quality data is extremely useful but sparse. To exploit this utility without excessive resource expenditure, we adopt a multi-phase training paradigm that implements selective repetition.
Without specification, each data sample appears exactly once within each phase, but high-quality samples may appear in multiple phases.
This ensures that higher-quality data samples are repeated more frequently. We employ a five-phase training pipeline, which limits repetition such that the overall benefit remains similar to that observed in one-pass training regimes~\citep{muennighoff2023data_constrained,yan2025largerdatasetsrepeatedmore}. (\cref{sec:multi_phase_mixture_subsec})

\paragraph{Multi-Domain Curriculum Training.}
In addition to strategic repetition, we integrate a data curriculum within training phases 3, 4, and 5. This curriculum ensures a stable data mixture across different datasets while sorting data samples in ascending order of their quality metrics within each dataset. Datasets without explicit quality labels are simply shuffled. This means that more important and high-quality samples are presented to the model in the latter training steps. To make full use of the benefit of the data curriculum, we adopt a moderate Learning Rate (LR) decay and apply model averaging over the last several checkpoints, following recent findings~\citep{luo2025learningratedecaywastes}. (\cref{sec:multi_dataset_curriculum_subsec})

\paragraph{Pushing the Frontier of Fully Open-Source Models.} 
\sys lies in the frontier of fully open-source models, achieving superior performance at comparable model scales. 
Most notably, our model demonstrates remarkable capabilities in three focused domains: Chinese language understanding, mathematical reasoning, and code generation. 
As illustrated in \Cref{tab:lang_math_code}, \sys substantially outperforms Gemma2-2B despite operating at a similar parameter scale. 
Beyond these targeted capabilities, our model also exhibits competitive performance in reasoning and knowledge-intensive tasks, approaching or matching the performance of larger models such as Gemma2-2B and Yulan-Mini-2.4B. 
When accounting for non-embedding parameters alone (only 1.4B), our model's efficiency advantages become even more pronounced, as demonstrated in \Cref{fig:performance_comparison_on_nonembedding}. (\cref{sec:eval})

The rest of this report is organized as follows: In \Cref{sec:architecture_stability}, we will discuss how to stabilize training on FP16-only hardware through architecture design. In \Cref{sec:data_benchmark}, we will introduce our quantile benchmarking approach to deepen our understanding of how various score metrics reflect the data inherent in different feature dimensions. In \Cref{sec:multi_phase_curriculum}, we will discuss two approaches to leverage high-quality data in our training: selective repetition and quality-based curriculum. Then in \Cref{sec:eval}, we will report our evaluation settings and results, positioning \sys in the fully-open and open-weight models. Additionally, \Cref{sec:non_embedding_comparison} shows model performance comparison relative to non-embedding parameters; \Cref{sec:quantile_benchmark_results} lists quantile benchmarking results; \Cref{sec:dataset_used} lists all used datasets along with license details; \Cref{sec:data_mixture_details} lists dataset mixture details in all phases; \Cref{sec:experiment_setting_details} presents the implementation details and experiment settings in our training and small-scale experiments. \Cref{sec:model_perf_full} shows the full table for model performance comparison.

\section{Architecture Design and Training Stability}
\label{sec:architecture_stability}

\begin{figure}[htbp]
    \centering
    \begin{subfigure}[t]{0.48\linewidth}
        \centering
        \includegraphics[width=\linewidth]{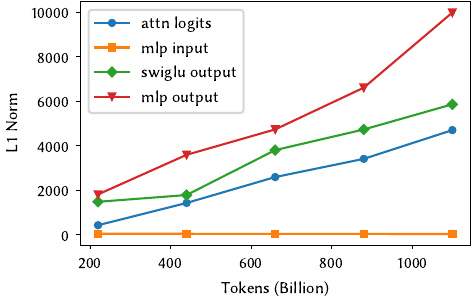}
        \caption{Activation statistics of the baseline architecture}
        \label{fig:2B_v3}
    \end{subfigure}
    \hfill
    \begin{subfigure}[t]{0.48\linewidth}
        \centering
        \includegraphics[width=\linewidth]{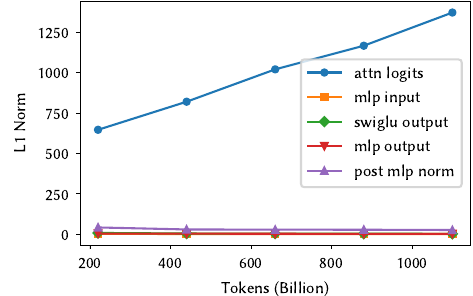}
        \caption{Activation statistics after applying Sandwich Normalization and Logits Soft-capping}
        \label{fig:2B_v4}
    \end{subfigure}
    
    \caption{Comparison of internal activation magnitudes before and after architectural optimization. The experiment is conducted with a 3B model.}
    \label{fig:main_figure_label}
\end{figure}

\sys is trained on Huawei Ascend 910A accelerators, which are similar to NVIDIA V100s in supporting only FP16 precision. However, FP16 has a limited dynamic numerical range, which introduces overflow risks when model parameters or activations grow too large. To keep training stable, we first identify the activations that are most likely to overflow and then introduce structural changes that keep their values within safe bounds.

Following the standard Llama architecture, the model uses SwiGLU~\cite{dauphin2017swiglu}, RMSNorm~\cite{zhang2019rmsnorm}, and RoPE~\cite{su2024rope}. We adopt mixed precision training, where operators that need higher precision, such as Softmax and RMSNorm, run in FP32, and the remaining computations run in FP16. Despite this setup, training on large and diverse datasets, including code and mathematics, still leads to strong numerical instability. As shown in \Cref{fig:2B_v3}, most instability comes from two places: the attention logits and the activations after the SwiGLU function in the MLP layers. In practice, the maximum activation values grow without control. They exceed $10{,}000$ after processing one trillion tokens, which is close to the FP16 upper limit. As a result, the dynamic loss scaler decreases its scaling factor to avoid overflow. This drop pushes many gradients below the FP16 minimum representable value, which causes underflow. The gradients then become inaccurate, harming convergence and sometimes causing training to fail.

To solve these issues, we use Logits Soft-Capping~\cite{bello2017softcapping} and Sandwich Normalization~\cite{ding2021cogview}. This follows the design choices of Gemma 2~\cite{morgane2024gemma2}. These techniques place strict bounds on activation values. As shown in \Cref{fig:2B_v4}, soft-capping reduces the L1 norm of attention logits by about an order of magnitude. At the same time, sandwich normalization reduces the accumulation of large values in residual connections and keeps the L1 norm of MLP activations within a safe range. To further improve stability, we set the weight decay to $0.1$, apply soft-capping to the final output logits, and replace the soft-capping inside each attention layer with QK-Norm~\cite{henry2020qknorm}. The full configuration of \sys is listed in \Cref{tab:training_config} and the implementation details are discussed in \Cref{sec:implement_architecture}.

\section{Data Benchmarking and Preprocessing}
\label{sec:data_benchmark}

There are many open-source pretraining datasets across various data domains, especially for English, Code, and Math~\cite{olmo20252olmo2furious,smol2,yiwen-etal-2025-yulan,DCLM,su2025nemotroncctransformingcommoncrawl,stack-v2,zhou2025megamath}.
However, constructing a high-quality pretraining corpus remains a non-trivial task due to two primary challenges. First, it is challenging to measure the quality of diverse datasets and determine the optimal strategy for selecting and mixing data from heterogeneous sources.
Second, preprocessing these datasets is both resource-intensive and technically complex.
Given the large scale of pretraining data and complex operations like deduplication, the preprocessing pipeline incurs substantial computational overhead and engineering complexity.

To mitigate these issues, (1) we propose to benchmark primary datasets (e.g., DCLM-Baseline~\cite{DCLM}, Fineweb-Edu~\cite{fineweb}) by quantiles of quality scores. We train reference models over the data subsets around a series of quantiles of quality scores, and then analyze how the resulting benchmark performance varies with data distribution, which is reflected by quality scores. (2) We develop a user-friendly Spark-based data preprocessing framework to efficiently process large-scale pretraining datasets. Moreover, we exploit the Chukonu~\cite{chukonu} framework to reduce the preprocess overhead. These explorations on data dimensions lay a solid foundation for our training and future work.

\subsection{Benchmarking Dataset By Quality-Score Quantiles}\label{sec:data_benchmarking_subsec}

\begin{figure}[ht]
    \centering
    \includegraphics[width=0.85\linewidth]{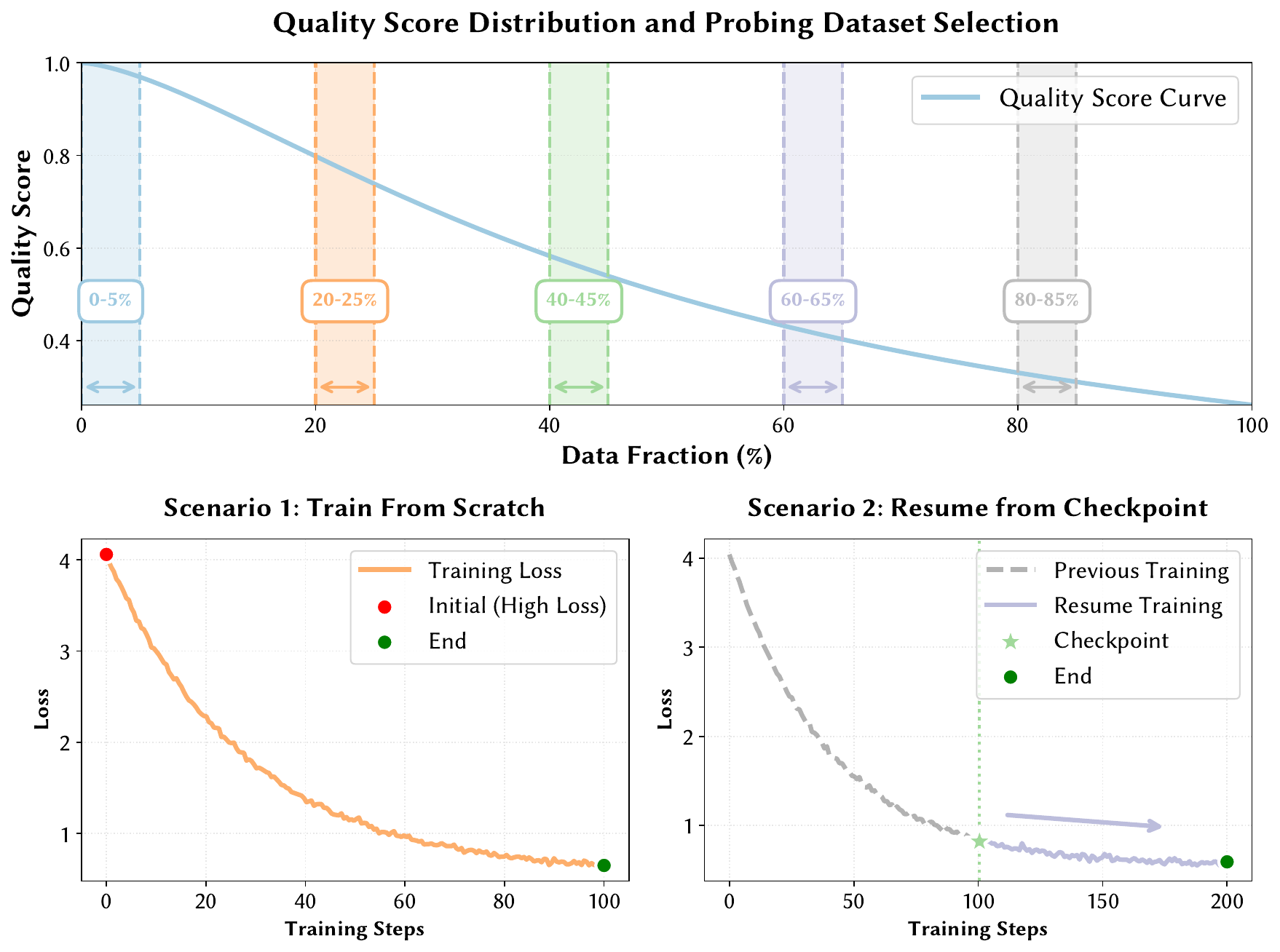}
    \caption{\textbf{Illustration of the Quantile Benchmarking Process.}
    (1) Given a series of target quantiles (e.g., 0\%, 20\%, 40\%, 60\%, 80\%), we select a data chunk around each target quantile as a probing dataset. 
    (2) A small-scale reference model is then evaluated on each probing dataset under two settings: training from scratch, or resuming from a checkpoint for continual training.}
    \label{fig:quantile_benchmark_process}
\end{figure}

\paragraph{Background and Motivation.} Most open-source datasets have been through a preprocessing pipeline, which primarily incorporates steps of quality scoring and data filtering by score. These score labels are typically released for these open-source datasets. Therefore, we can select a subset hypothesized to be of high quality based on sample quality scores. However, samples between different datasets are hard to compare, considering heterogeneous quality metrics. When scorers are available, it is possible to score both datasets. But as more datasets and quality metrics are included, it is hard to scale up and judge by multiple quality metrics. DCLM~\cite{DCLM} proposes to benchmark datasets or quality scores by filtering and feeding datasets into a standard series of models across scales. However, when facing a practical pretraining setup, the top-k filtering and multi-scale benchmarking can be challenging: the cost of full benchmarking is prohibitive, and we need to ablate over different filtering ratios to balance quality and quantity of the filtered dataset. 

\subsubsection{Quantile Benchmarking Methodology}

\textit{Quantile Data Benchmarking} refers to the systematic evaluation of dataset quality by training reference models on data subsets stratified by quality score percentiles, enabling empirical comparison of dataset characteristics across different quality ranges. Instead of relying solely on top-$k$ filtering, we design a small-scale evaluation process across a range of quality score quantiles to benchmark dataset quality. In practice, preprocessed open-source datasets typically provide a quality metric for each data sample. These quality scores reflect specific characteristics of the data and can vary significantly. It is commonly expected that using higher-quality data samples in training should lead to better model performance. Motivated by this intuition, we propose a straightforward approach, which has not been systematically explored in prior work.

For a target dataset, we first determine a series of quality score quantiles, such as top 0\%, 10\%, 20\%, $\dots$, 80\%. At each quantile, we select a fixed-size subset of the dataset. In our implementation, starting from the data sample ranked at the top 10\% in terms of quality score, we expand the subset by including the next 10B tokens of lower-quality samples to form the probing dataset. We then train a small-scale reference model (e.g., 0.5B parameters) on each of these probing datasets. Finally, we evaluate the resulting models on a set of target benchmarks to record their performances. We refer to this process of evaluating datasets across different quality quantiles as \textit{quantile benchmarking}.

Given the computational cost of training multiple reference models on different probing datasets, we typically apply quantile benchmarking to dominant datasets, such as DCLM-Baseline~\cite{DCLM} and FineWeb-Edu~\cite{fineweb} in the English domain, and FineWeb-Edu-Chinese-V2.1~\cite{fineweb-edu-chinese} in the Chinese domain. Moreover, we measure the utility of each probing dataset in two scenarios: (1) training the reference model from scratch, and (2) continuing training from pretrained checkpoints. Evaluating both scenarios provides a more comprehensive understanding of the target dataset. \Cref{fig:quantile_benchmark_process} illustrates the overall quantile benchmarking process, including quantile data selection and benchmarking experiments under both scenarios. 

While quantile benchmarking introduces additional computational overhead, this cost is marginal compared to the resources required for full pretraining.
For instance, as illustrated in \Cref{fig:quantile_benchmark_results}, we benchmark the DCLM-Baseline across 5 quantiles, ranging from 0\% to 60\% at 15\% intervals.
Although the entire deduplicated DCLM-Baseline comprises 609B tokens (as reported in \Cref{tab:phase1_stats}), our benchmarking process consumes only 8.4B tokens per run, totaling 42B tokens (approximately 6.9\% of the dataset).
Furthermore, because we utilize a reference model of at most 0.6B parameters, the total computational budget constitutes only 2\% of the cost required to train a 2B target model over the full DCLM-Baseline in a single pass, and represents less than 0.6\% of the total pretraining budget of \sys (Explained in \Cref{sec:referece_quantile_benchmarking}).
Given that quantile benchmarking is performed only once per target dataset to guide all subsequent selection decisions, the actual amortized cost is negligible.
Beyond efficiency, these experiments offer significantly higher granularity than training a reference model over a subset via top-$k$ filtering, which merely reflects the average quality of data above a specific threshold rather than the distribution within specific bands.
Consequently, we strategically target this benchmarking at leading datasets, such as DCLM-Baseline, Fineweb-Edu, and Fineweb-Edu-Chinese-V2.1, where the approach exhibits the most favorable benefit-cost ratio.

\subsubsection{Benchmarking Results and Analysis}
Based on our quantile experiment results, we compare models trained on different dataset partitions across various benchmarks. These quantile-based comparisons provide deeper insights into the characteristics of target datasets and offer guidance for data selection and mixing strategies.

\begin{figure}
    \begin{center}
        \begin{subfigure}[b]{0.48\linewidth}
            \centering
            \includegraphics[width=\linewidth]{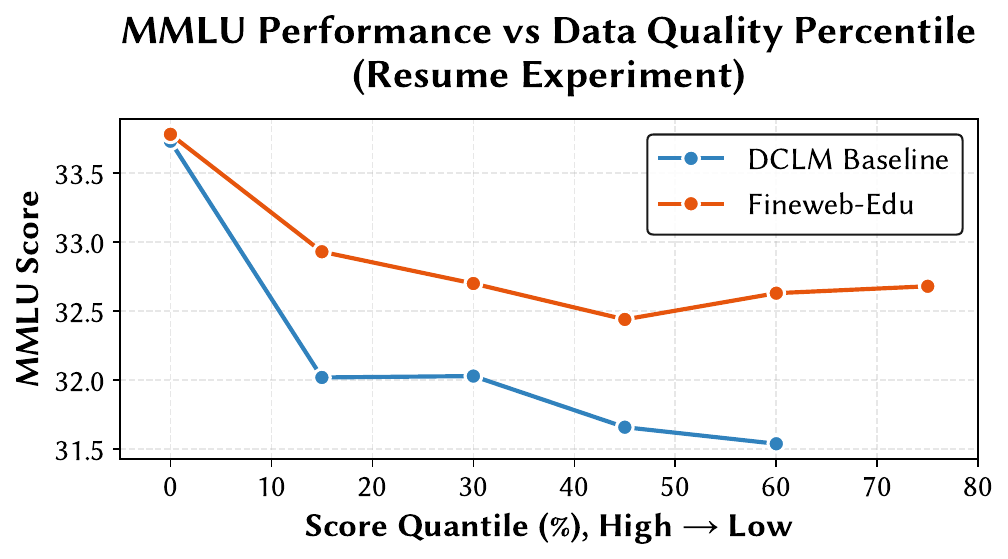}
        \end{subfigure}
        \begin{subfigure}[b]{0.48\linewidth}
            \centering
            \includegraphics[width=\linewidth]{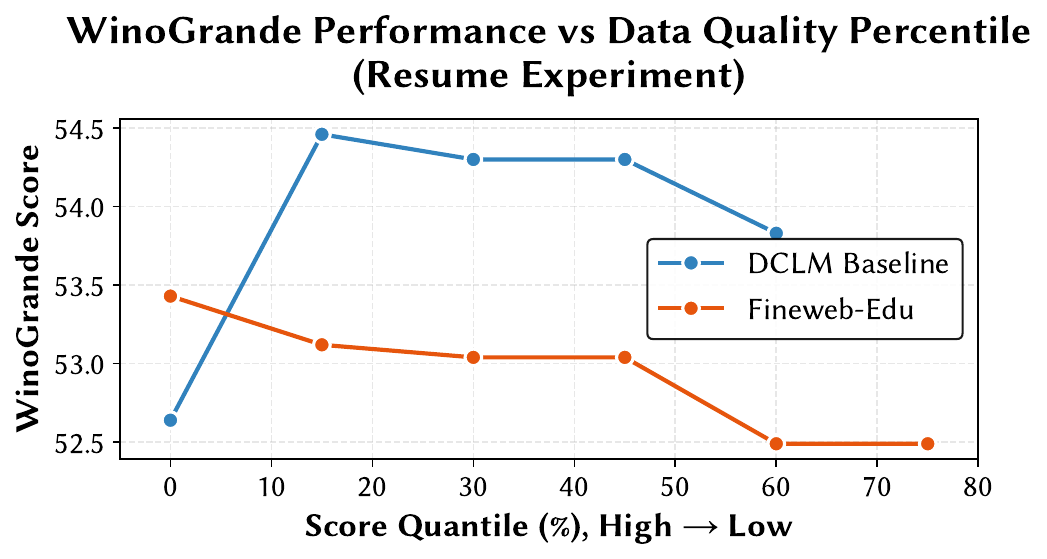}
        \end{subfigure}
    \end{center}
    \caption{Representative results showing task-dependent dataset characteristics: FineWeb-Edu excels on know\-ledge-intensive tasks (MMLU) while DCLM-Baseline performs better on commonsense reasoning (WinoGrande).}
    \label{fig:quantile_benchmark_results}
\end{figure}

As an illustrative example, we present quantile experiments on both Fineweb-Edu and DCLM-Baseline, offering a complementary perspective to previous analyses~\cite{su2025nemotroncctransformingcommoncrawl,wettig2025organizewebconstructingdomains}. The representative comparison sees \Cref{fig:quantile_benchmark_results} and full comparison results can refer to \Cref{fig:quantile_benchmark_understanding,fig:quantile_benchmarks_knowledge}. Our investigation aims to deepen the understanding of these representative open-source datasets and identify their key differences and commonalities, which we summarize as follows:

\begin{enumerate}
    \item[(1)] \textbf{Task-dependent dataset superiority.} Fineweb-Edu generally demonstrates superior performance on academic and encyclopedic benchmarks, including MMLU~\cite{mmlu} and Common Sense QA (CSQA)~\cite{commonsenseqa}, as well as reading comprehension tasks like BoolQ~\cite{clark-etal-2019-boolq}. In contrast, DCLM-Baseline exhibits slight advantages on situated commonsense reasoning, such as PIQA~\cite{piqa}, Social IQa~\cite{socialiqa}, HellaSwag~\cite{hellaswag}, and WinoGrande~\cite{winogrande}. This divergence suggests that FineWeb-Edu excels in tasks requiring more structural knowledge and formal semantics, while DCLM may benefit tasks relying on intuitive scenario-based reasoning. Representative comparisons are illustrated in \Cref{fig:quantile_benchmark_results} where Fineweb-Edu induces better results on MMLU while DCLM-Baseline outperforms on WinoGrande. More comprehensive results are presented in \Cref{fig:quantile_benchmarks_knowledge,fig:quantile_benchmark_understanding}, which respectively highlight academic knowledge and formal reasoning, and situated commonsense reasoning.
    \item[(2)] \textbf{Substantial within-dataset heterogeneity.} Data quality varies considerably within individual datasets. For instance, in the continual training (resume) scenario, DCLM-Baseline exhibits a 2\% performance difference on MMLU between the top 0\% and top 60\% quantiles, while showing an even more pronounced 8\% variation on ARC-Easy across the same quality range, as shown in \Cref{tab:reasoning_knowledge}. This substantial heterogeneity underscores the importance of quality-aware data selection and training strategies.
    
    \item[(3)] \textbf{Consistency across training scenarios.} The relative superiority relationships between datasets remain largely consistent across both continual training (resume) and from-scratch (run) scenarios, as demonstrated in \Cref{fig:quantile_benchmark_understanding,fig:quantile_benchmarks_knowledge}. However, we observe occasional deviations in specific quantile ranges, suggesting that training dynamics may influence relative dataset effectiveness.
    
    \item[(4)] \textbf{Non-monotonic quality-performance relationships.} Benchmark performance does not necessarily increase monotonically with quality scores. As shown in \Cref{fig:quantile_benchmark_understanding,fig:quantile_benchmarks_knowledge}, increasing quality scores, measured by the fineweb-edu classifier for Fineweb-Edu and the FastText score for DCLM-Baseline, can paradoxically lead to decreased performance on HellaSwag and PIQA. This finding calls into question the universal applicability of quality metrics employed in current leading open-source datasets, and highlights the task-specific nature of data quality assessment.
\end{enumerate}

In summary, our quantile experiments reveal that (i) datasets exhibit substantial internal heterogeneity, and (ii) the relative superiority of both datasets and their quality partitions is highly dependent on the target capability of interest. 
Quality assessment is task-dependent and context-specific, necessitating careful consideration when comparing datasets across different capability dimensions. More details of implementation and discussions are presented in \Cref{sec:referece_quantile_benchmarking}.

\subsubsection{Implications for Data Selection}
These findings inform our data mixing and training strategies in the following ways:

\begin{enumerate}
    \item[(1)] \textbf{Curriculum learning with selective repetition.} Beyond the conventional practice of filtering low-quality data, we propose strategically scheduling high-quality data partitions toward later training stages (curriculum learning) while applying higher repetition rates to these partitions compared to lower-quality data (selective multi-epoch training). This approach leverages within-dataset quality variation to enhance training efficiency (details in \Cref{sec:multi_phase_curriculum}).
    
    \item[(2)] \textbf{Benchmark-guided mixing ratios.} Given a representative benchmark aligned with a target capability, such as MMLU for knowledge-intensive tasks, quantile comparisons can guide inter-dataset mixing ratios. For example, as illustrated in the left panel of \Cref{fig:quantile_benchmark_results}, the entire Fineweb-Edu dataset exhibits performance roughly comparable to the top 30\% partition of DCLM-Baseline on MMLU, suggesting appropriate relative sampling rates for knowledge-focused pretraining. 
    Concretely, in Phase 2 (\Cref{tab:phase2_stats}), we use the whole Fineweb-Edu dataset while using only the top 33.4\% DCLM-Baseline dataset. In the latter phase, the relative ratio of DCLM-Baseline is further pulled down, shown in \Cref{tab:phase3_stats,tab:phase4_stats,tab:phase5_stats}. The heuristic mixing ratio design lays a foundation for more strategies introduced in \Cref{sec:multi_phase_curriculum}.
\end{enumerate}

We acknowledge that the current analysis remains primarily qualitative and coarse-grained. More fine-grained, quantitative frameworks for dataset comparison and mixing ratio optimization represent promising directions for future research.

\subsection{Data Processing Framework}
\label{sec:data_preprocessing_framework}

To address the challenges of data processing, our data processing framework is designed to satisfy three critical requirements:
\begin{enumerate}
    \item \textbf{Reproducibility:} Given that the training dataset of \sys is composed of various open-source datasets, the framework should be able to reconstruct the exact dataset from these original sources with a configuration file.
    \item \textbf{Usability and Scalability:} The framework should support various operations like filtering, deduplication and mixing. Furthermore, this framework should scale to large clusters without additional engineer efforts.
    \item \textbf{High Performance:} To handle hundreds of terabytes of data, the framework must be optimized to reduce computation overhead.
\end{enumerate}

To meet these demands, we developed \textbf{\dataframework}, a distributed data processing framework built on Spark~\cite{zaharia2012spark}.
\dataframework adopts a tree-structured processing pipeline design.
The leaf nodes represent the raw open-source datasets, while internal nodes represent processing operators like filters and samplers.
The root node generates the final mixed training dataset.
With this design, the entire processing pipeline, including dataset sources and operator parameters, can be defined with a YAML configuration file.
This ensures strict reproducibility, enabling researchers to reconstruct the exact training corpus from raw datasets simply by applying the configuration.

As \dataframework is built on Spark, it inherits the programming flexibility and scalability.
We utilize the powerful Spark RDD API to develop complex data processing operators, and rely on the Spark Engine for distributed processing, resource management, and fault tolerance.
This design allows \dataframework to process over 100 TB of data across large-scale clusters with minimal engineering efforts.

Despite Spark's scalability, the overhead of JVM-based execution can become a bottleneck for compute-intensive tasks. To address this, we integrated the Chukonu~\cite{chukonu} framework, utilizing its C++ interface to refactor certain performance-critical operators.
By conducting computations with native C++, we accelerates the processing procedure.
For instance, the optimized MinHash deduplication operator is approximately $2.5\times$ faster than the Spark implementation.

\section{Multi-Phase Multi-Dataset Curriculum Training}
\label{sec:multi_phase_curriculum}

Data quality heterogeneity within datasets, as revealed in Section~\ref{sec:data_benchmark}, presents both opportunities and challenges for model training. High-quality samples can significantly enhance model capabilities more efficiently than average-quality data, yet they typically constitute only a small fraction of the overall dataset. To leverage this heterogeneity, we propose two principles: (1) progressive exposure, where higher-quality data appears in later training phases, and (2) strategic repetition, where high-quality partitions will be repeated more. In implementation, we design data curriculum at both phase and instance levels, and repeat data across different phases, thereby amplifying the impact of valuable training samples and improving overall data utilization efficiency. The multi-phase training practice is also adopted in other open-source model training pipelines~\citep {smol2,hu2024yulan}.

\begin{table*}[t]
\centering
\caption{Performance Comparison: Selective Repetition and Curriculum Learning Strategies.}
\label{tab:curriculum_comparison}
\resizebox{\textwidth}{!}{
\begin{tabular}{@{}llcccccccccc@{}}
\toprule
\textbf{Method} & \textbf{Retain} & \textbf{MMLU} & \textbf{ARC-c} & \textbf{ARC-e} & \textbf{CSQA} & \textbf{OBQA} & \textbf{PIQA} & \textbf{SIQA} & \textbf{Wino.} & \textbf{Avg.} & \textbf{Core} \\
\midrule
Uniform& 100\% & 30.77 & 42.14 & 61.05 & 50.86 & 45.20 & 72.42 & 45.75 & 56.27 & 50.56 & 46.21 \\
\midrule
\rowcolor{ours}
CMA& 100\% & 31.68 & \textbf{41.47} & \textbf{61.93} & \textbf{52.50} & \textbf{46.00} & 71.71 & 45.39 & 57.22 & \textbf{50.99} & \textbf{46.89} \\
\midrule
Filter\&Repeat& 13.8\% & \textbf{32.99} & 35.79 & 61.75 & 46.03 & 42.00 & 71.71 & 44.37 & 56.35 & 48.87 & 44.14 \\
\rowcolor{fullyopen}
Filter\&Repeat& 33.4\% & 32.44 & 41.14 & \textbf{61.93} & 51.11 & 43.80 & 72.09 & 45.34 & \textbf{58.80} & 50.83 & 46.65 \\
Filter\&Repeat& 77.4\% & 31.68 & 38.46 & 60.70 & \textbf{52.50} & 45.00 & \textbf{72.52} & \textbf{45.80} & 57.22 & 50.49 & 45.83 \\
\bottomrule
\end{tabular}
}
\end{table*}

\subsection{Multi-Phase Domain Mixture and Quality-Based Selective Repetition}\label{sec:multi_phase_mixture_subsec}

\begin{figure}
    \centering
    \begin{subfigure}[b]{0.49\textwidth}
        \centering
        \includegraphics[width=\linewidth]{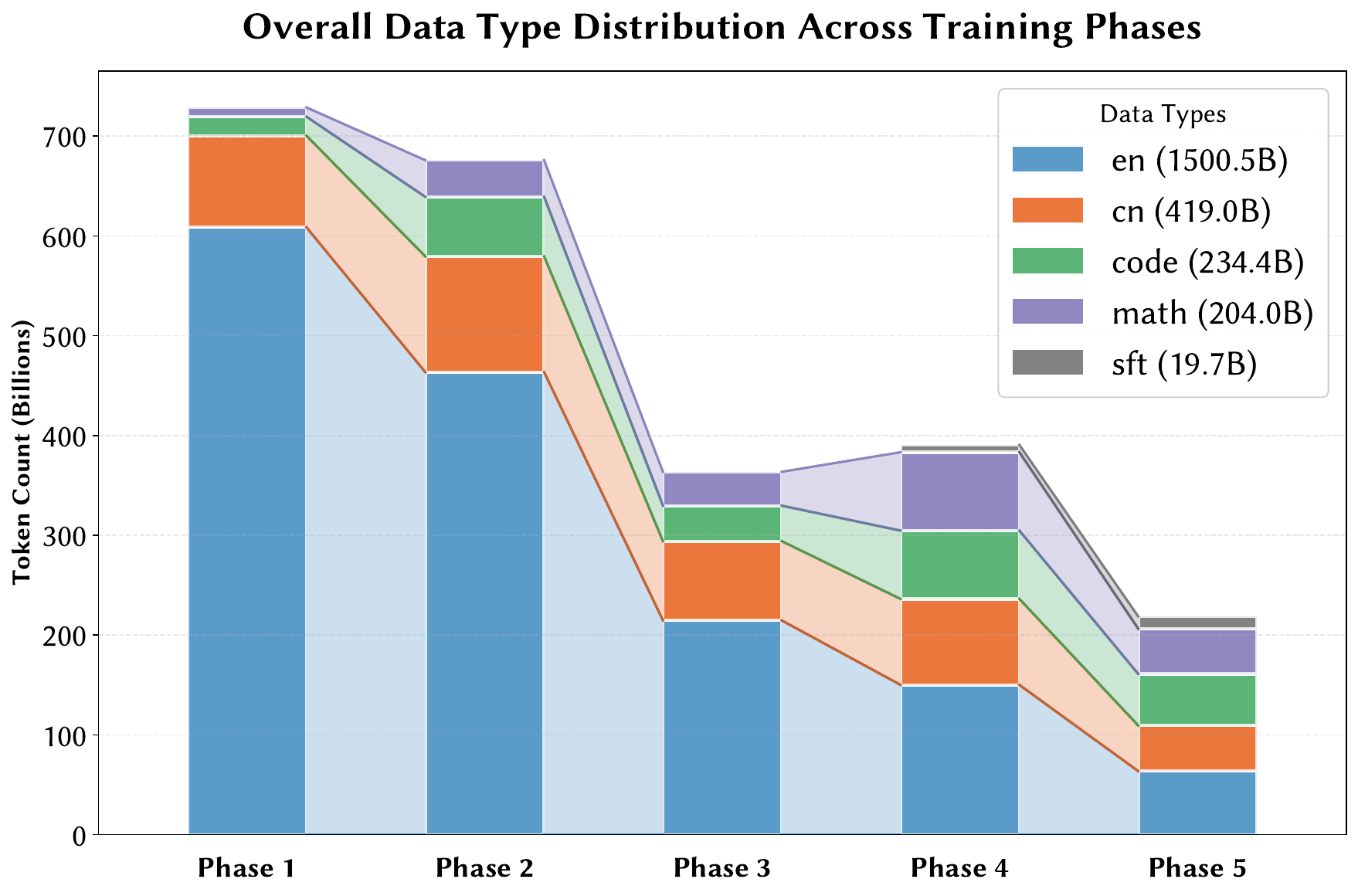}
        \caption{Phase-wise data mixture transitions}
    \end{subfigure}%
    \hfill
    \begin{subfigure}[b]{0.49\textwidth}
        \centering
        \includegraphics[width=\linewidth]{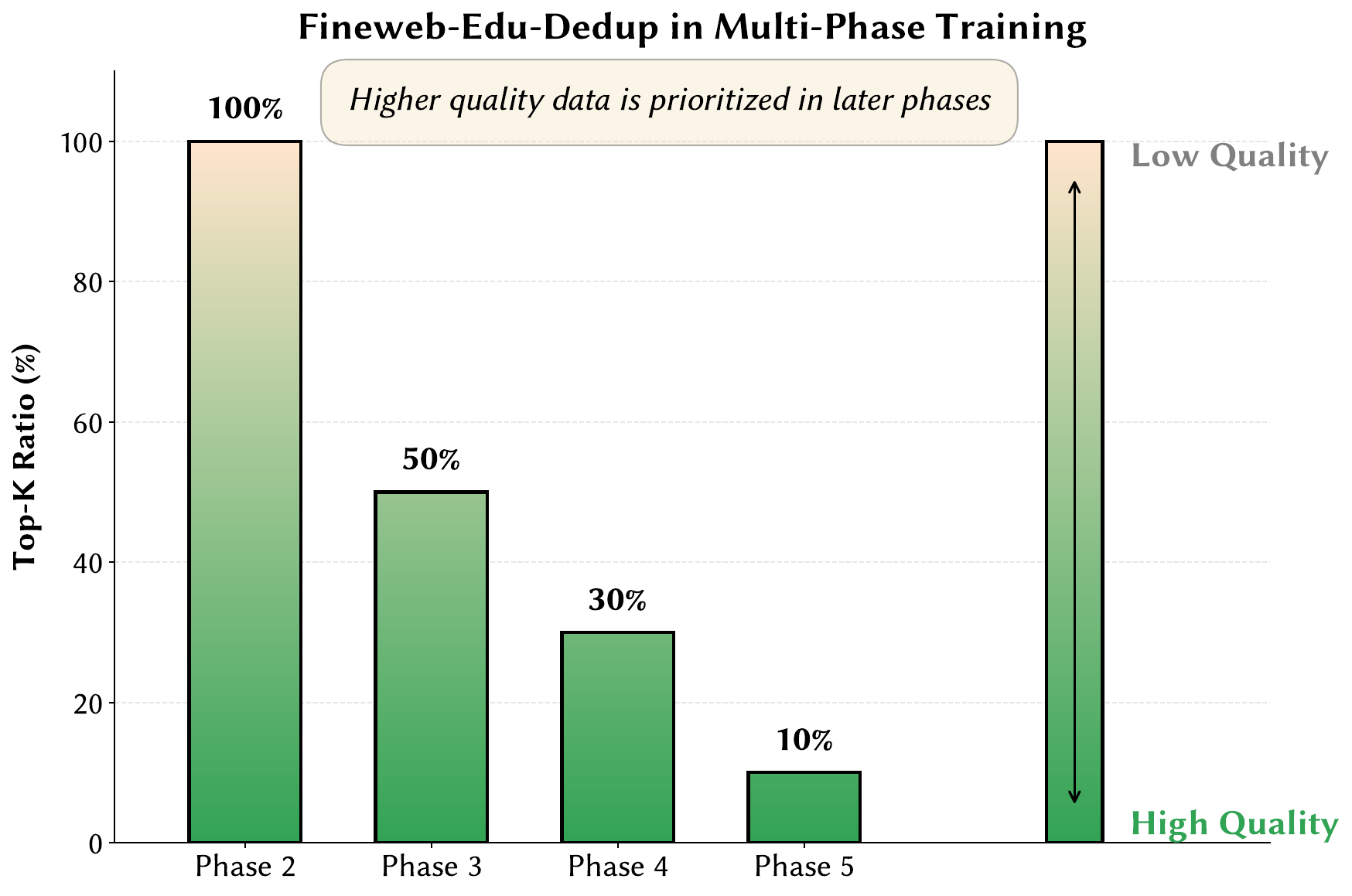}
        \caption{Phase-wise top-k ratio for Fineweb-Edu}
    \end{subfigure}%
    \caption{Five training phases of \sys. Latter phases keep more refined data samples.}
    \label{fig:data_mixture}
\end{figure}

We structure the training process into five distinct phases with progressive data mixtures, as illustrated in \Cref{fig:data_mixture}. This phased approach incorporates two perspectives of curriculum strategies: domain-level progression and quality-based selection and repetition.

First, we implement a domain-level curriculum by gradually increasing the proportion of Chinese, code, and mathematical datasets in later phases, while introducing supervised fine-tuning data in the final two phases. The phase-wise mixture transitions are visualized in \Cref{fig:data_mixture}. To keep training stability, we maintain English content above 30\% while limiting Chinese, code, and mathematical content each below 30\%. The specific domain mixtures are detailed in \Cref{tab:phase1_stats,tab:phase2_stats,tab:phase3_stats,tab:phase4_stats,tab:phase5_stats}.

Second, we apply quality-based filtering within each domain during later phases, retaining only high-scoring partitions based on available quality metrics. Specifically, one dataset can occur in multiple phases rather than appear only once in the whole training process. However, in each phase, each data sample mostly occurs only once. Moreover, instead of repeating the whole dataset, we mostly keep only the high-quality partition and progressively decrease top-k retention ratios across phases, effectively increasing average data quality, for datasets with a quality metric. 
For example, as shown in \Cref{fig:data_mixture}, we use the entire Fineweb-Edu dataset in phase two, then retain only 50\%, 30\%, and 10\% of top-quality samples in subsequent phases. This selective repetition strategy ensures that the top 10\% samples are exposed to the model four times throughout training, while lower-quality samples appear only once.

This selective repetition serves two primary purposes. On the one hand, we experimentally find that mildly repeating a high-quality portion can attain better training efficiency than one-pass training. We validate this approach using a 1.5B Qwen2.5 model trained on 30B tokens from a DCLM-Baseline shard. As shown in Table~\ref{tab:curriculum_comparison}, retaining 33.4\% of top-quality samples for three epochs outperforms one-pass training, demonstrating the efficacy of strategic repetition. More repetition on the aggressive filtered subset can improve even more on benchmarks like MMLU, but can impede other general capabilities. Experimental details are in \Cref {sec:reference_repetition_curriculum}. A similar experiment is conducted in D4~\citep{tirumala2025d4} while we step forward to explore the effect of filtering and repetition on different capability dimensions, rather than on merely PPL or loss perspective. On the other hand, repetition compensates for aggressive deduplication, as high-quality content naturally occurs more frequently in the internet and can also serve as an indicator of data quality. Prior research indicates that mild multi-epoch training (under four repetitions) preserves sample utility, with larger datasets tolerating more repetition~\citep{yan2025largerdatasetsrepeatedmore,muennighoff2023data_constrained}.

\subsection{Multi-Dataset Data Curriculum}\label{sec:multi_dataset_curriculum_subsec}

Beyond phase-level adjustments, we construct instance-level curriculum learning within each phase. To fully take advantage of the data curriculum, we adopt the technique of Curriculum Model Average (CMA)~\citep{luo2025learningratedecaywastes}, which adopts appropriate learning rate scheduling and model averaging in curriculum-based pretraining. As demonstrated in \Cref{tab:curriculum_comparison}, CMA outperforms uniform sampling in our 1.5B model experiments. We discuss the small-scale reference experiment in \Cref{sec:reference_repetition_curriculum} in detail.

\begin{figure}[t!]
    \centering
    \includegraphics[width=1.0\linewidth]{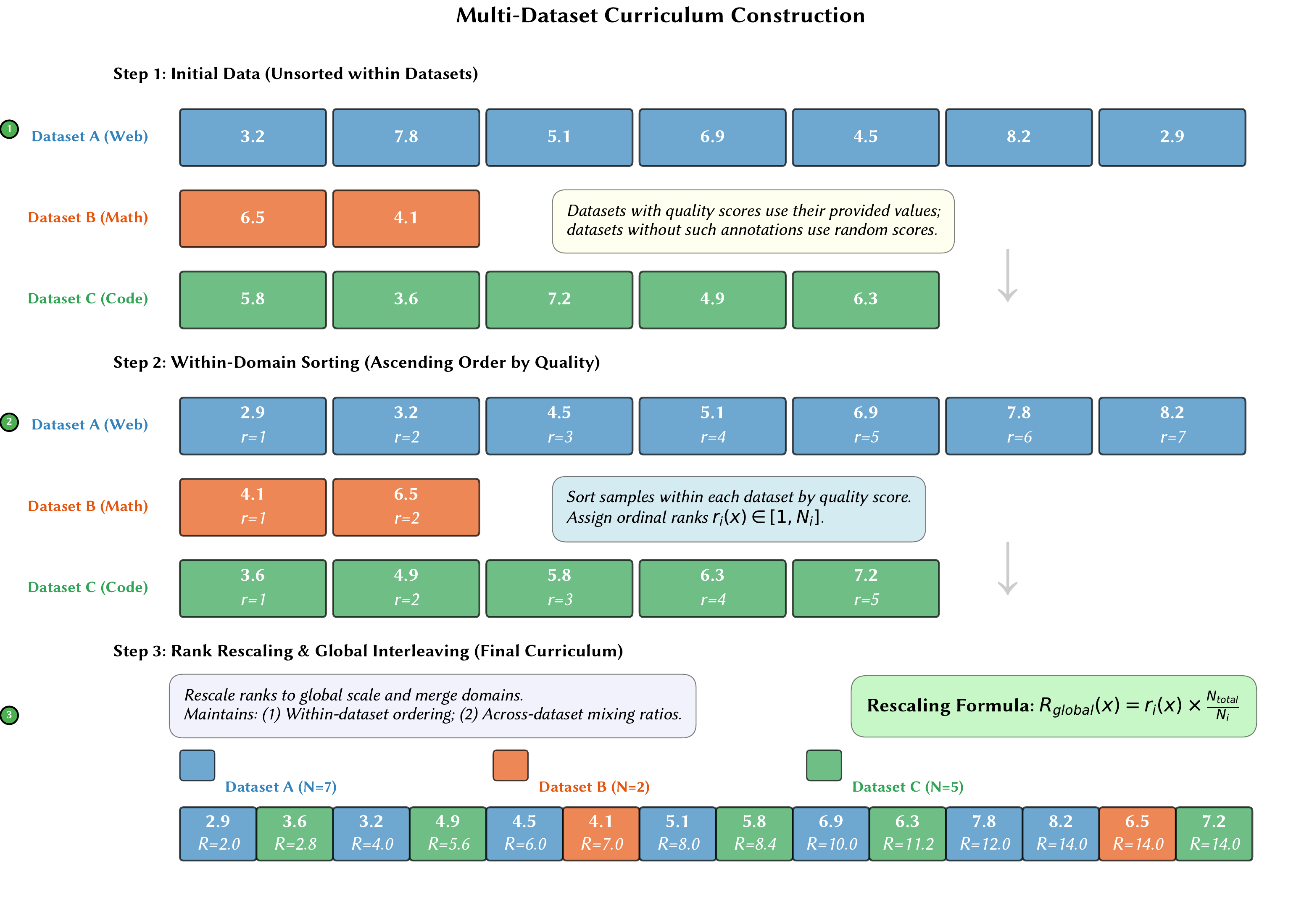}
    \caption{Multi-Dataset Curriculum Construction Process.}
    \label{fig:multi_domain_curriculum}
\end{figure}

However, the pretraining dataset mostly consists of data samples from various source corpora and constructing a multi-dataset curriculum will present new challenges. Different datasets may employ distinct quality metrics or lack them entirely. To address this, we propose the three-step procedure outlined in \Cref{alg:multi-domain-curriculum} and \Cref{fig:multi_domain_curriculum}:

\begin{enumerate}
    \item \textbf{Within-Dataset Ranking}: Samples within each dataset are independently sorted using dataset-specific quality metrics in ascending order. For datasets without quality labels, we assign random scores uniformly distributed in $[0,1]$ to enable shuffling within the curriculum framework.
    \item \textbf{Rank Rescaling}: Dataset-specific ranks are normalized to a global scale using:
    \[
    R_{\text{global}}(x_A) = r_A \times \frac{N_{\text{total}}}{N_A}
    \]
    where $r_A$ is the within-dataset rank, $N_A$ is the dataset sample count, and $N_{\text{total}}$ is the total sample count.
    
    \item \textbf{Global Interleaving}: All samples are merged and sorted by their rescaled global ranks.
\end{enumerate}

This algorithm ensures: (1) preservation of within-dataset quality ordering or shuffling the dataset without quality labels, (2) proportional interleaving across datasets according to mixture ratios, and (3) maintenance of stable dataset mixtures throughout training. The mixture of datasets is heuristically designed with reference to the quantile benchmarking results in \Cref{sec:data_benchmark}.

\begin{algorithm}[t]
\caption{Multi-Dataset Curriculum Construction}
\label{alg:multi-domain-curriculum}
\begin{algorithmic}[1]
\Require Datasets $D_1, D_2, \dots, D_k$ with their specific quality metrics
\Ensure Multi-dataset curriculum dataset
\State $\displaystyle N_{\text{total}} \gets \sum_{i=1}^k |D_i|$ \Comment{Compute total sample count}
\For{each dataset $D_i$}
    \State (Optionally) Add a random number for the dataset without a quality label
    \State Sort $D_i$ by dataset-specific quality metric (ascending) \Comment{Within-dataset ranking}
    \State Assign ordinal ranks $r_i(x) \in [1, |D_i|]$ to each sample $x \in D_i$
    \State Compute rescaled ranks: $R(x) \gets r_i(x) \times \frac{N_{\text{total}}}{|D_i|}$ for all $x \in D_i$
\EndFor
\State $\displaystyle U \gets \bigcup_{i=1}^k D_i$ \Comment{Combine all datasets}
\State Sort $U$ by rescaled rank $R(x)$ in ascending order \Comment{Global interleaving}
\State \Return sorted $U$
\end{algorithmic}
\end{algorithm}

In practice, we implement this multi-dataset curriculum in the final three training phases to avoid low-quality samples being sorted together and fed to an immature model, which can result in instability. We set the final learning rate to $6\times 10^{-4}$ and average the last six checkpoints, as detailed in Section~\ref{sec:training_configuration}.

\subsection{Pretraining Configuration and Implementation}
\label{sec:training_configuration}

\paragraph{Model Architecture.} Our 2B-parameter model architecture primarily refers to Qwen3-1.7B~\citep{qwen3} with modifications for training stability. We untie word embeddings to reduce communication overhead, resulting in 1.4B non-embedding parameters and 0.6B embedding parameters. To ensure FP16 training stability, we incorporate QK-norm, sandwich norm, and soft capping (detailed in \Cref{sec:architecture_stability}). Complete architectural specifications are provided in \Cref{tab:training_config}.

\paragraph{Training Hyperparameters.} We train with a context length of 4096 and a batch size of 2048. We run small-scale experiments with a 1.5B model on 30B tokens at a batch size of 512, and then extrapolate roughly an optimal peak learning rate of $5\times 10^{-3}$ via square-root scaling~\citep{malladi2022sqrtrule}. We employ a Warmup-Stable-Decay schedule~\citep{hu2024minicpmunveilingpotentialsmall} and reduce the peak learning rate from $5\times 10^{-3}$ to $3\times 10^{-3}$ after the first phase to mitigate the instability caused by data distribution shift effects. The learning rate remains constant through phases 2--4, then decays to $6\times 10^{-4}$ in phase 5 to accommodate the multi-dataset curriculum~\citep{luo2025learningratedecaywastes}. We average the final eight checkpoints (saved every 3.36B tokens) to reduce variance from insufficient learning rate decay, with model averaging details provided in \Cref{sec:model_average}.

\begin{figure}
    \centering
    \includegraphics[width=0.9\linewidth]{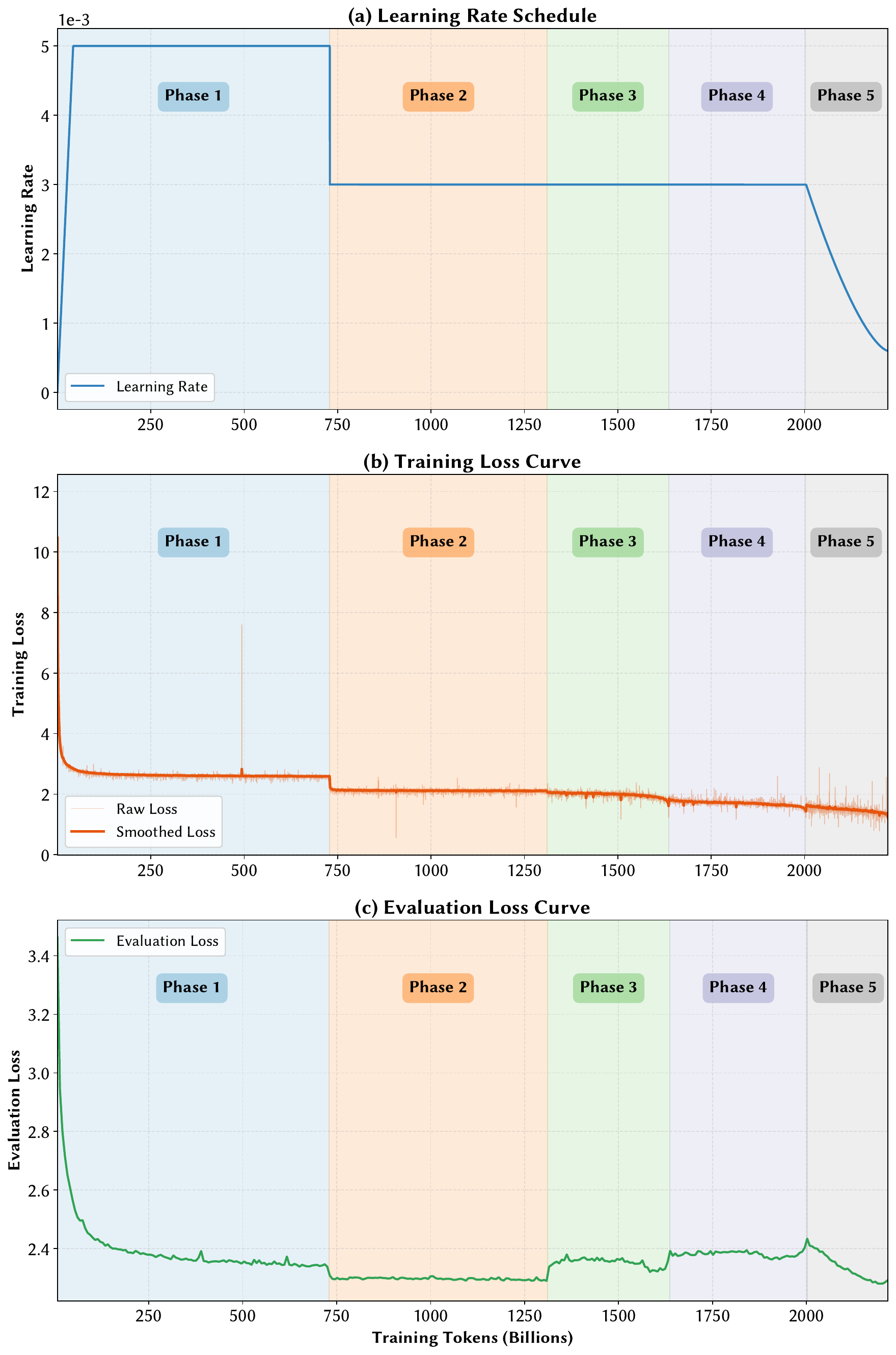}
    \caption{Learning Rate Schedule, Training Loss, and Validation Loss Curves.}
    \label{fig:lr_loss}
\end{figure}

\paragraph{Training Curve Analysis.}
\Cref{fig:lr_loss} presents learning rate, training loss, and validation loss trajectories. Two key observations emerge:

\begin{enumerate}
    \item Training loss exhibits non-standard decay patterns due to three factors: (1) the learning rate reductions between phase 1 and phase 2 induce abrupt loss drop at the phase transition, (2) in later phases,  we introduce increased low-perplexity code and mathematical content, which results in continually decreasing loss, and (3) the quality-based curricula import more high-quality data in latter steps within phases 3--5, which accelerates convergence of loss, followed by slight increases at phase transitions.
    
    \item Validation loss on a DCLM-Baseline subset shows similar phase-transition drops but anomalous increases during phases 3--4. These increases likely reflect domain misalignment between the validation set (primarily English text) and training data (increasing code and mathematical content). Each phase ends with accelerated validation loss decay (benefiting from high-quality data) followed by sharp increases (probably from distribution shifts by curriculum).
\end{enumerate}

These patterns suggest that more diverse validation sets would better track training progress. Future work should consider more gradual domain transitions and increased phase counts for improved training stability.

\section{Evaluation}\label{sec:eval}

We conduct extensive evaluations on \sys and compare it against both open-weight and fully open models. Our \sys advances the frontier of fully open models and further narrows the gap between fully open models and leading open-weight models.

\subsection{Evaluation Setup}

\subsubsection{Baseline Models}
We evaluate \sys against a comprehensive set of state-of-the-art baseline models with comparable parameter counts. These baselines are categorized into two distinct groups: \textbf{open-weight models}, where model weights are public but training data or details may remain proprietary, and \textbf{fully-open models}, where the architecture, weights, training code, and datasets are all publicly released.\footnote{All evaluated models are base (pretrained) checkpoints without instruction tuning. To maintain consistency, we standardize naming conventions by omitting suffixes (e.g., using simplified names for Qwen3) to denote base models throughout this paper.}

\paragraph{Open-weight models.}
\begin{itemize}
    \item \textit{Qwen2-1.5B}~\cite{qwen2}: A 1.5B-parameter decoder-only transformer trained on large-scale multilingual and code data. It delivers robust performance in general language understanding, coding, and reasoning while facilitating efficient deployment.
    \item \textit{Qwen2.5 series}~\cite{qwen2025qwen25technicalreport}: We select Qwen2.5-1.5B and Qwen2.5-3B, dense foundation models that refine the Qwen2.5 architecture. These models feature an improved tokenizer and offer enhanced capabilities in knowledge, coding, and mathematics within a compact form factor suitable for edge applications.
    \item \textit{Qwen3 series}~\cite{qwen3}: We include Qwen3-0.6B, Qwen3-1.7B, and Qwen3-4B. These are small-scale base models that support long contexts and ``thinking'' modes, providing competitive abilities in general tasks, mathematics, and coding.
    \item \textit{Gemma2-2B}~\cite{morgane2024gemma2}: The smallest member of Google's Gemma 2 family, this model is distilled from larger counterparts. It was trained on 2 trillion tokens from diverse sources, including web documents, code, and scientific articles.
    \item \textit{Llama3.2 series}~\cite{llama3.2}: We utilize Llama-3.2-1B and Llama-3.2-3B, multilingual text-only models distilled and pruned from larger Llama variants. They support extended context windows (128k) and tool-calling, targeting privacy-preserving on-device inference.
\end{itemize}

\paragraph{Fully-open models.}
\begin{itemize}
    \item \textit{SmolLM2-1.7B}~\cite{smol2}: Developed by Hugging Face, this model utilizes the Llama 2 architecture with a GPT-2 tokenizer (vocabulary size 49,152). It was trained on 256 H100 GPUs.
    \item \textit{SmolLM3-3B}~\cite{smollm3}: A compact, fully-open model trained on 11T tokens using data-centric recipes. It features a 128k context window utilizing NoPE and YaRN, offering state-of-the-art performance for its size class with multilingual support.
    \item \textit{OLMo2-1B}~\cite{olmo20252olmo2furious}: The smallest model in the OLMo2 family (specifically OLMo2-0425-1B), trained on the OLMo-mix corpus. Its full release of code, checkpoints, logs, and training details enables rigorous scientific inquiry into compute-efficient training at the 1B-parameter scale.
    \item \textit{YuLan-Mini}~\cite{yiwen-etal-2025-yulan}: A 2.4B-parameter model pre-trained on approximately 1.08T tokens. By combining curated data pipelines with robust optimization and annealing strategies, it achieves top-tier performance among similarly sized models, particularly in mathematics and coding.
\end{itemize}

\subsubsection{Benchmarks}
Our evaluation encompasses four primary domains: mathematics, coding, Chinese language processing, and general reasoning \& knowledge. We selected representative benchmarks for each domain as follows:
\begin{itemize}
    \item \textit{Math}: We utilize GSM8K~\cite{gsm8k} and MATH~\cite{MATH}. Together, these datasets cover the spectrum from grade-school arithmetic to advanced competition-style problems, providing a comprehensive assessment of symbolic and multi-step reasoning.
    \item \textit{Coding}: We adopt MBPP~\cite{mbpp} and HumanEval~\cite{humaneval} to evaluate code generation via unit testing. This approach directly measures the model's ability to synthesize executable and logically coherent programs. Specifically, we use the sanitized subset of MBPP, which refines problem descriptions and test cases to minimize ambiguity.
    \item \textit{Chinese}: To assess knowledge and reasoning within a Chinese linguistic context, we employ CMMLU~\cite{cmmlu} and C-Eval~\cite{ceval}, widely adopted benchmarks spanning diverse academic and professional subjects.
    \item \textit{Reasoning \& Knowledge}: For general English-language knowledge and reasoning, we include a suite of eight datasets: MMLU~\cite{mmlu}, HellaSwag~\cite{hellaswag}, Common Sense QA (CSQA)~\cite{commonsenseqa}, BoolQ~\cite{clark-etal-2019-boolq}, PIQA~\cite{piqa}, SocialIQA~\cite{socialiqa}, WinoGrande~\cite{winogrande}, and ARC~\cite{arc}. These benchmarks cover a broad range of expert knowledge, reading comprehension, and commonsense reasoning scenarios.
\end{itemize}

\subsubsection{Implementation Details}
We conduct our evaluation using the OpenCompass framework~\cite{2023opencompass}, a comprehensive platform for large model evaluation. For mathematics and coding benchmarks, which typically require open-ended generation, we evaluate models in \textit{generation mode}. Conversely, for benchmarks in other domains, we employ \textit{perplexity-based (PPL) evaluation}. Following the OLMES protocol~\cite{gu2025olmes}, PPL tasks are assessed under both multiple-choice formulation (MCF) and completion formulation (CF), with the superior score reported as the final result.

\subsection{Evaluation Results}

The performance of \sys and baseline models is summarized in \Cref{tab:lang_math_code,tab:reasoning_knowledge}, with a comprehensive comparison provided in \Cref{tab:model_comparison_full}.

\begin{table}[!htbp]
\centering
\caption{Core Capabilities: Chinese, Math, and Code.}
\label{tab:lang_math_code}
\resizebox{\textwidth}{!}{%
\begin{threeparttable}
\begin{tabular}{lccccccccc}
\toprule
\multirow{3}{*}{\textbf{Model Name}} & \multirow{3}{*}{\textbf{Params}} & \multicolumn{2}{c}{\textbf{Chinese}} & \multicolumn{2}{c}{\textbf{Math}} & \multicolumn{2}{c}{\textbf{Code}} & \multirow{3}{*}{\textbf{Avg}}\\
\cmidrule(lr){3-4} \cmidrule(lr){5-6} \cmidrule(lr){7-8}
 &  & C-Eval & CMMLU  & GSM8K & MATH & sanitized-MBPP & HumanEval & \\
 &  & 5 shot & 5 shot & 4 shot & 4 shot & 3 shot & 3 shot & \\
\midrule

\multicolumn{9}{l}{\textit{\textbf{Open-Weight SOTA}}} \\
\rowcolor{openweight} Qwen2-1.5B & 1.5B & 71.29 & 70.62 & 58.50$^{*}$ & 21.70$^{*}$ & 50.58 & 31.10$^{*}$ & 50.63\\
\rowcolor{openweight} Qwen2.5-1.5B & 1.5B & 68.63 & 68.01  & 68.50$^{*}$ & 35.00$^{*}$ & 58.37 & 37.20$^{*}$ & 55.95\\
\rowcolor{openweight} Qwen2.5-3B & 3B & 74.65 & 73.92  & 79.10$^{*}$ & 42.60$^{*}$ & 66.54 & 42.10$^{*}$ & 63.15\\
\rowcolor{openweight} Qwen3-0.6B & 0.6B & 57.03 & 52.36 & 59.59$^{*}$ & 32.44$^{*}$ & 51.75	 & 29.88 & 47.18 \\
\rowcolor{openweight} Qwen3-1.7B & 1.7B & 66.70 & 66.55  & 75.44$^{*}$ & 43.5$^{*}$ & 64.20 & 52.44 & 61.47\\
\rowcolor{openweight} Qwen3-4B & 4B & 78.5 & 77.01 & 87.79$^{*}$ & 54.10$^{*}$ & 74.32 & 62.20 & 72.32\\
\rowcolor{openweight} Gemma2-2B & 2B & 41.35 & 39.63 & 23.90$^{*}$ & 15.00$^{*}$ & 38.91 & 17.70$^{*}$ & 29.42\\
\rowcolor{openweight} Llama-3.2-1B & 1B & 29.82 & 31.03 & 44.40$^{*}$ & 30.60$^{*}$ & 34.63 & 18.90 & 31.56\\
\rowcolor{openweight} Llama-3.2-3B & 3B & 45.67 & 44.33 & 77.70$^{*}$ & 48.00$^{*}$ & 49.42 & 29.88 & 49.17\\

\multicolumn{9}{l}{\textit{\textbf{Fully-Open SOTA}}} \\
\rowcolor{fullyopen} SmolLM2-1.7B & 1.7B & 35.06 & 34.03 & 31.10$^{*}$ & 11.60$^{*}$ & 49.42 & 22.60$^{*}$ & 30.64\\
\rowcolor{fullyopen} OLMo-2-0425-1B & 1B & 30.53 & 28.62 & 68.30$^{*}$ & 20.70$^{*}$ & 15.56 & 6.71 & 28.40\\
\rowcolor{fullyopen} YuLan-Mini-2.4B & 2.4B & 52.32 & 48.14 & 66.65$^{*}$ & 27.12 & 62.26 & 61.60$^{*}$ & 53.02\\
\rowcolor{fullyopen} SmolLM3-3B & 3B & 50.84 & 49.35 & 67.63$^{*}$ & 46.10$^{*}$	& 62.26 & 39.63 & 52.64\\

\multicolumn{9}{l}{\textit{\textbf{Ours}}} \\
\rowcolor{ours} PCMind-2.1-Kaiyuan-2B & 2B & 46.30 & 49.25 & 51.33 & 30.34 & 56.42 & 42.68 & 46.05\\

\bottomrule
\end{tabular}%

\begin{tablenotes}
\item[*] This score is cited from the corresponding official report or paper.
\end{tablenotes}
\end{threeparttable}
}
\end{table}

\textbf{Core Capabilities: Math, Code, and Chinese.} 
In \Cref{tab:lang_math_code}, we focus on three specialized capabilities of the model: mathematics, coding, and Chinese language\footnote{For generation tasks (math and code), we report official results for baseline models where available, as exact reproduction can be challenging.}. \sys achieves an average score of \texttt{46.05} across these seven benchmarks. It outperforms fully-open models of similar scale, such as SmolLM2-1.7B and OLMo-2-0425-1B, and remains competitive with larger models like YuLan-Mini-2.4B and SmolLM3-3B, despite a smaller parameter count.
Specifically, on Chinese benchmarks (C-Eval and CMMLU), \sys scores \texttt{46.30} and \texttt{49.25}, respectively—markedly higher than SmolLM2-1.7B and OLMo-2-0425-1B, and approaching the performance of the larger SmolLM3-3B. On mathematics, \sys achieves \texttt{51.33} on GSM8K, substantially surpassing SmolLM2-1.7B, and scores \texttt{30.34} on MATH, outperforming YuLan-Mini-2.4B (\texttt{27.12}). Similarly, in code generation, our model reaches \texttt{42.68} on HumanEval, exceeding both SmolLM3-3B (\texttt{39.63}) and Qwen2.5-3B (\texttt{42.10}). The results demonstrate \sys offers a superior trade-off between accuracy and model size in critical domains.

\textbf{Reasoning and Knowledge.}
\Cref{tab:reasoning_knowledge} presents performance on nine reasoning and knowledge benchmarks. \sys achieves an average score of \texttt{67.74}, placing it firmly within the competitive range for its size class. Within the fully-open category, our model surpasses SmolLM2-1.7B (+1.69 average) and OLMo-2-0425-1B (+5.68 average), while effectively matching the larger YuLan-Mini-2.4B (\texttt{67.50}). Although the larger SmolLM3-3B attains a higher average (\texttt{72.60}), \sys significantly narrows the gap to the state-of-the-art for fully-open models at this scale.
When compared to open-weight models, \sys achieves an average score of \texttt{67.74}, approaching Gemma2-2B (\texttt{69.16}) despite using comparable training data. 
Larger open-weight models like Qwen3-4B maintain a substantial lead (\texttt{81.84}), which is expected given their significantly larger scale and training resources.

\begin{table}[htbp]
\centering
\caption{Reasoning and Knowledge Capabilities.}
\label{tab:reasoning_knowledge}
\resizebox{\textwidth}{!}{%
\begin{tabular}{lcccccccccccc}
\toprule
\multirow{3}{*}{\textbf{Model Name}} & \multirow{3}{*}{\textbf{Params}} & \multicolumn{9}{c}{\textbf{Reasoning \& Knowledge}} & \multirow{3}{*}{\textbf{Avg}}\\
\cmidrule(lr){3-11}
 &  & MMLU & ARC-C & ARC-E & BoolQ & CSQA & HSwag & PIQA & SocIQ & Wino & \\
&  & 5 shot & 5 shot & 5 shot & 5 shot & 5 shot & 5 shot & 5 shot & 5 shot & 5 shot &\\
\midrule

\multicolumn{11}{l}{\textit{\textbf{Open-Weight SOTA}}} \\
\rowcolor{openweight} Qwen2-1.5B & 1.5B & 56.36 & 70.17 & 83.60 & 71.90 & 70.52 & 60.77 & 75.73 & 63.46 & 59.83 & 68.04\\
\rowcolor{openweight} Qwen2.5-1.5B & 1.5B & 61.56 & 79.32 & 90.48 & 76.39 & 75.10 & 64.18 & 76.17 & 64.94 & 59.67 & 71.98 \\
\rowcolor{openweight} Qwen2.5-3B & 3B & 66.86 & 86.44 & 92.59 & 83.88 & 76.09 & 73.85 & 81.45 & 69.40 & 63.69 & 77.14\\
\rowcolor{openweight} Qwen3-0.6B & 0.6B & 55.09 & 68.14 & 84.48 & 69.05 & 61.18 & 48.51 & 69.97 & 61.51 & 55.64 & 63.73 \\
\rowcolor{openweight} Qwen3-1.7B & 1.7B & 65.35 & 80.34 & 91.89 & 79.82 & 74.61 & 60.76 & 77.20 & 68.58 & 59.27 & 73.09\\
\rowcolor{openweight} Qwen3-4B & 4B & 75.78 & 89.83 & 97.53 & 86.09 & 81.9 & 79.46 & 84.98 & 75.59 & 65.43 & 81.84\\
\rowcolor{openweight} Gemma2-2B & 2B & 55.20 & 66.44 & 82.54 & 72.42 & 69.45 & 66.20 & 78.89 & 65.92 & 65.35 & 69.16\\
\rowcolor{openweight} Llama-3.2-1B & 1B & 37.74 & 36.95 & 70.55 & 67.43 & 62.82 & 60.20 & 74.92 & 50.61 & 58.17 & 57.71\\
\rowcolor{openweight} Llama-3.2-3B & 3B & 57.87 & 72.20 & 83.95 & 76.73 & 70.35 & 71.06 & 79.05 & 64.33 & 64.09 & 71.07\\

\multicolumn{11}{l}{\textit{\textbf{Fully-Open SOTA}}} \\
\rowcolor{fullyopen} SmolLM2-1.7B & 1.7B & 51.99 & 59.66 & 82.72 & 69.85 & 67.16 & 65.30 & 78.51 & 60.18 & 59.12 & 66.05\\
\rowcolor{fullyopen} OLMo-2-0425-1B & 1B & 44.25 & 47.46 & 76.72 & 70.55 & 65.60 & 61.61 & 76.44 & 55.53 & 60.38 & 62.06\\
\rowcolor{fullyopen} YuLan-Mini-2.4B & 2.4B & 51.76 & 64.75 & 82.54 & 78.59 & 66.18 & 61.20 & 77.31 & 63.25 & 61.88 & 67.50\\
\rowcolor{fullyopen} SmolLM3-3B & 3B & 63.04 & 77.29 & 88.54 & 76.12 & 70.52 & 69.20 & 79.05 & 65.25 & 64.40 & 72.60\\

\multicolumn{11}{l}{\textit{\textbf{Ours}}} \\
\rowcolor{ours} PCMind-2.1-Kaiyuan-2B & 2B & 53.90 & 66.10 & 82.89 & 78.53 & 67.40 & 58.13 & 74.37 & 62.59 & 65.75 & 67.74\\

\bottomrule
\end{tabular}%

}
\end{table}

\paragraph{Discussion on Size and Performance Trade-offs.} The overall trade-off between model size and average benchmark performance is visualized in \Cref{fig:model_perf_comp}. The figure reveals that \sys lies beyond the current fully-open frontier: at comparable parameter counts, it clearly outperforms earlier fully-open models (e.g., OLMo-2-1B, SmolLM2-1.7B) and approaches the performance of the larger YuLan-Mini-2.4B. Moreover, if adhering to the convention of comparing non-embedding parameters to get rid of the vocabulary effect, our \sys can exhibit even more prominent advantages, as shown in \Cref{fig:performance_comparison_on_nonembedding}.
We compare different models according to non-embedding parameters in \Cref{sec:non_embedding_comparison}.

Furthermore, when compared to open-weight baselines of similar size, \sys demonstrates superior architectural efficiency. 
For instance, compared to Gemma2-2B, \sys utilizes fewer model parameters (1.4B non-embedding and 2B total parameters, versus Gemma2-2B's 2B non-embedding and 2.6B total parameters) while maintaining a similar token budget (2.2T tokens for\sys versus 2T for Gemma2-2B). 
Despite these reduced resource requirements, \sys achieves stronger performance on core capabilities (Chinese, Math, Code) and competitive reasoning scores, as shown in \Cref{tab:lang_math_code,tab:reasoning_knowledge,tab:model_comparison_full}. 
Although a performance gap remains compared to the Qwen series, likely attributable to their massive training data scale (e.g., 36T tokens), \sys occupies a favorable position in the size-performance landscape, offering a strong, fully-open alternative for resource-constrained environments.

\subsection{Non-Embedding Based Comparison}
\label{sec:non_embedding_comparison}

\begin{figure}[htbp]
    \centering
    \includegraphics[width=\linewidth]{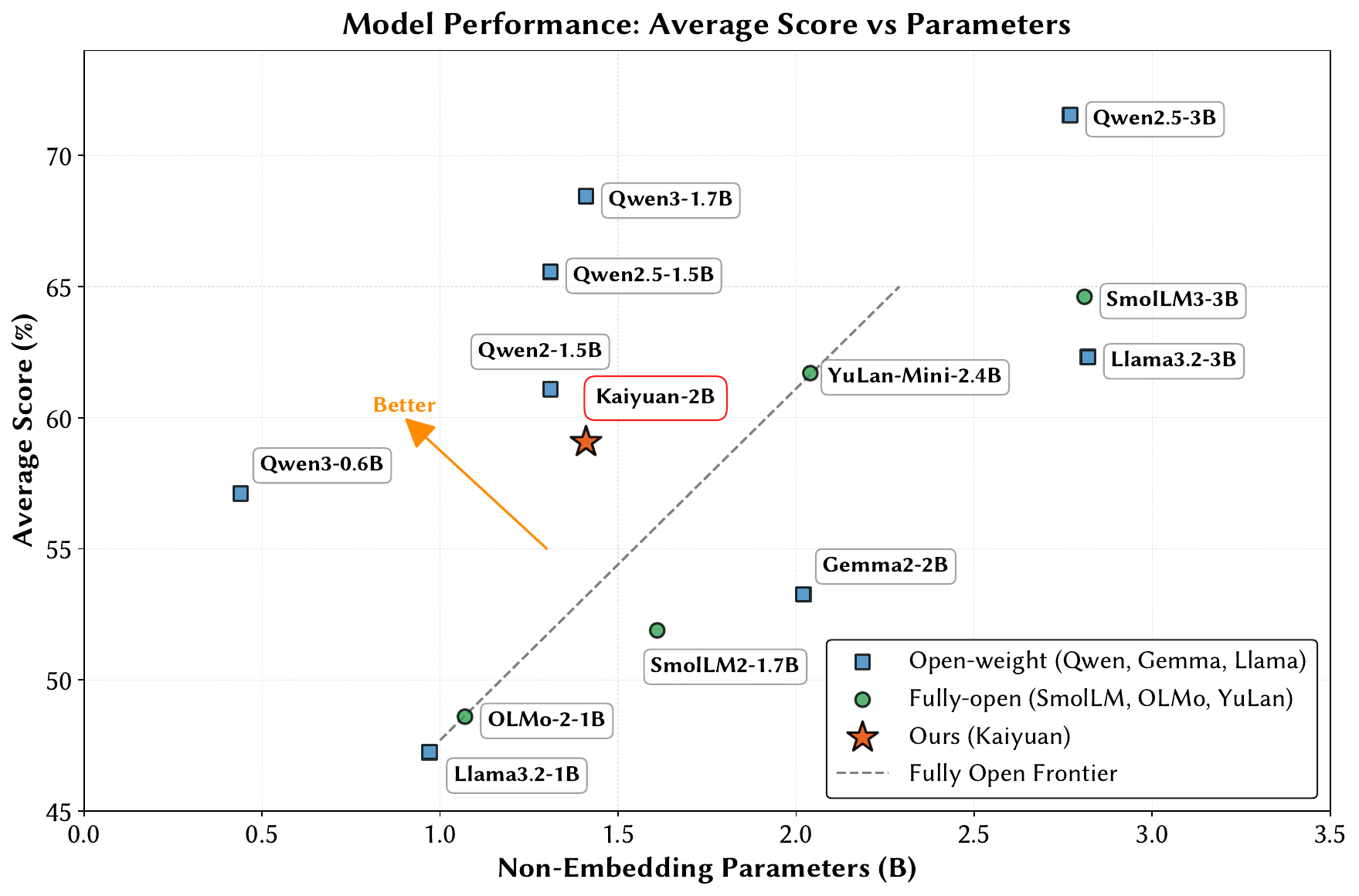}
    \caption{Model Performance Comparison over Non-embedding Parameters.}
    \label{fig:performance_comparison_on_nonembedding}
\end{figure}

In practice, the vocabulary sizes are different across different models, and embedding layers commonly account for relatively lower compute per parameter. We also note that the naming of different models has no consensus on using total parameters or non-embedding parameters in the model name. Hence, to conduct a more complete comparison, we also take statistics on both total parameters and non-embedding parameters of different open-weight and fully open-source models, and report the results in \Cref{tab:model_params_comparison}. In addition, taking non-embedding parameter as the X-axis, we report an additional comparison in \Cref{fig:performance_comparison_on_nonembedding}. We find that our model still excels the frontier of fully open-source models, and approaches close to leading open-weight models, like the Qwen series of a similar scale. Our advantage over other fully open-source models looks more prominent when taking account the non-embedding parameters.

\begin{table}[htbp]
\centering
\caption{Model Parameter Statistics Comparison.}
\label{tab:model_params_comparison}
\begin{tabular}{lcccc}
\toprule
\textbf{Model Name} & \textbf{Total} & \textbf{Embedding} & \textbf{Non-Embedding} & \textbf{Tied Embedding}\\
\midrule
\multicolumn{5}{l}{\textit{\textbf{SOTA Models}}} \\
\rowcolor{openweight} Qwen2-1.5B & 1.54B & 0.23B & 1.31B & TRUE\\
\rowcolor{openweight} Qwen2.5-1.5B & 1.54B & 0.23B & 1.31B & TRUE\\
\rowcolor{openweight} Qwen2.5-3B & 3.09B & 0.31B & 2.77B & TRUE\\
\rowcolor{openweight} Qwen3-0.6B-Base & 0.60B & 0.16B & 0.44B & TRUE\\
\rowcolor{openweight} Qwen3-1.7B-Base & 1.72B & 0.31B & 1.41B & TRUE\\
\rowcolor{openweight} Qwen3-4B-Base & 4.02B & 0.39B & 3.63B & TRUE\\
\rowcolor{openweight} Gemma-2-2B & 2.61B & 0.59B & 2.02B & TRUE\\
\rowcolor{openweight} Llama-3.2-1B & 1.24B & 0.26B & 0.97B & TRUE\\
\rowcolor{openweight} Llama-3.2-3B & 3.21B & 0.39B & 2.82B & TRUE\\
\multicolumn{5}{l}{\textit{\textbf{Fully-Open SOTA Models}}} \\
\rowcolor{fullyopen} SmolLM2-1.7B & 1.71B & 0.10B & 1.61B & TRUE\\
\rowcolor{fullyopen} OLMo-2-0425-1B & 1.48B & 0.41B & 1.07B & FALSE\\
\rowcolor{fullyopen} YuLan-Mini & 2.42B & 0.38B & 2.04B & FALSE\\
\rowcolor{fullyopen} SmolLM3-3B & 3.08B & 0.26B & 2.81B & TRUE\\
\multicolumn{5}{l}{\textit{\textbf{Ours}}} \\
\rowcolor{ours} PCMind-2.1-Kaiyuan-2B & 2.03B & 0.62B & 1.41B & FALSE\\
\bottomrule
\end{tabular}
\end{table}

\section{Conclusion}

The \sys project successfully demonstrates a systematic and resource-efficient approach to fully open-source LLM pretraining, providing concrete answers to the challenges of data heterogeneity and computational scarcity. Our core contributions include Quantile Data Benchmarking, Strategic Manual Repetition, and Multi-Domain Curriculum Training. Together, they represent a practical framework for the academic community to select and utilize public data effectively. By releasing the model checkpoint, the open-source data preprocessing framework, and the final pretraining dataset, we provide a complete, transparent recipe for high-quality LLM pretraining. We believe \sys is a valuable contribution that will facilitate further exploration and innovation in the open-source LLM ecosystem, pushing the frontier of what is achievable under limited resources.

\bibliographystyle{ACM-Reference-Format}
\bibliography{references, datasets}

\clearpage

\appendix

\begin{center}
\Large
Appendices
\end{center}

\section{Quality-Score Quantile Benchmarking}
\label{sec:quantile_benchmark_results}

\begin{figure}[htbp]
    \centering
    \begin{subfigure}[b]{0.49\linewidth}
    \includegraphics[width=\linewidth]{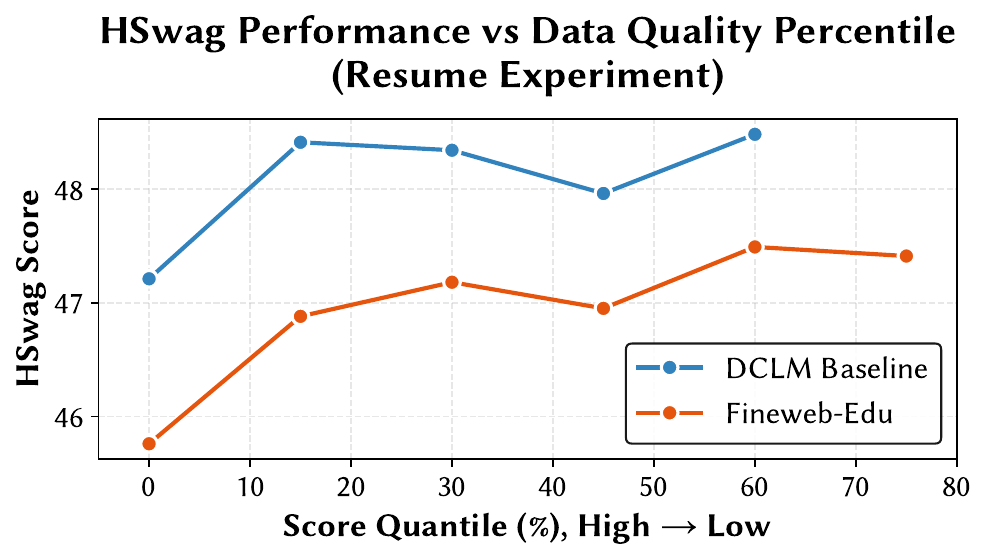} 
    \end{subfigure}
    \begin{subfigure}[b]{0.49\linewidth}
    \includegraphics[width=\linewidth]{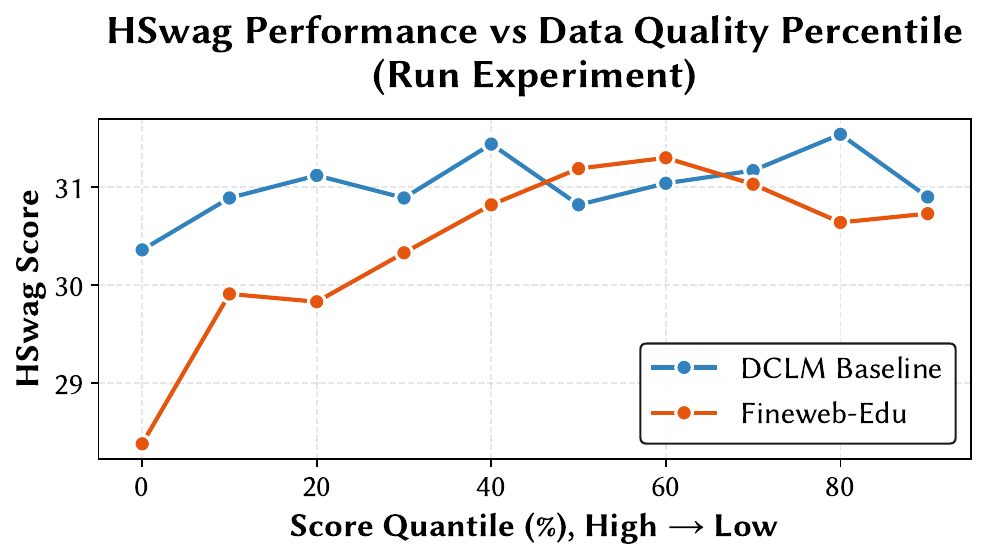} 
    \end{subfigure}
    \\
    \begin{subfigure}[b]{0.49\linewidth}
    \includegraphics[width=\linewidth]{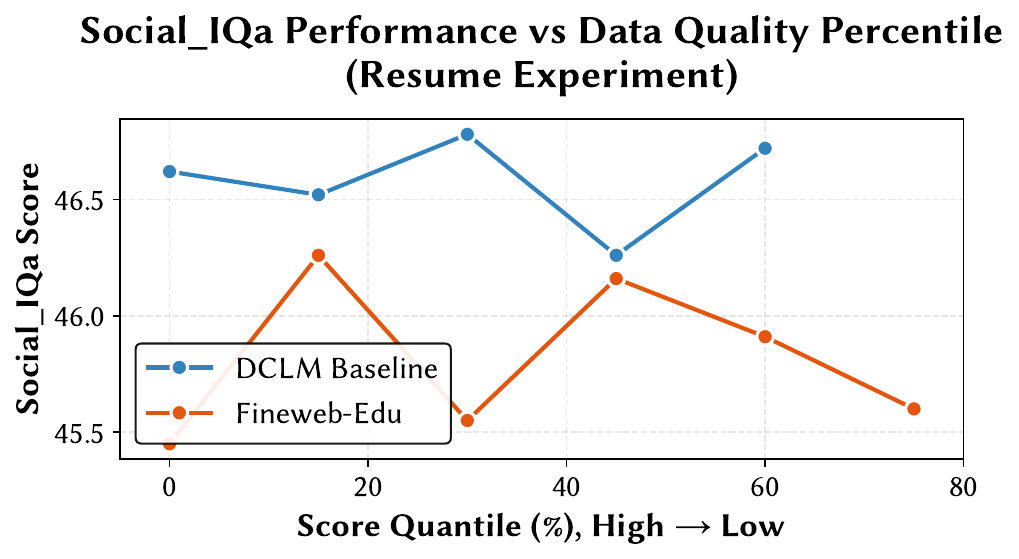} 
    \end{subfigure}
    \begin{subfigure}[b]{0.49\linewidth}
    \includegraphics[width=\linewidth]{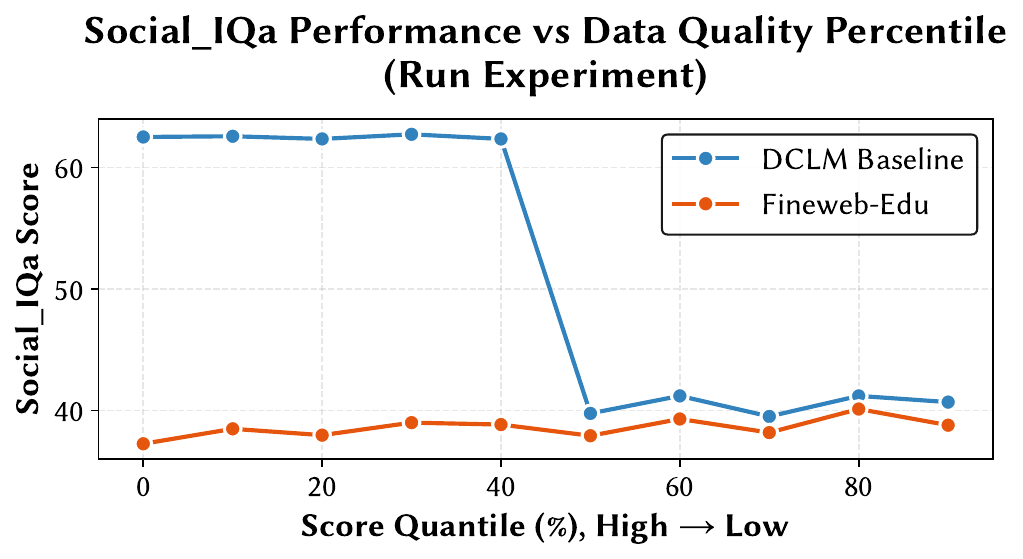} 
    \end{subfigure}
    \\
    \begin{subfigure}[b]{0.49\linewidth}
    \includegraphics[width=\linewidth]{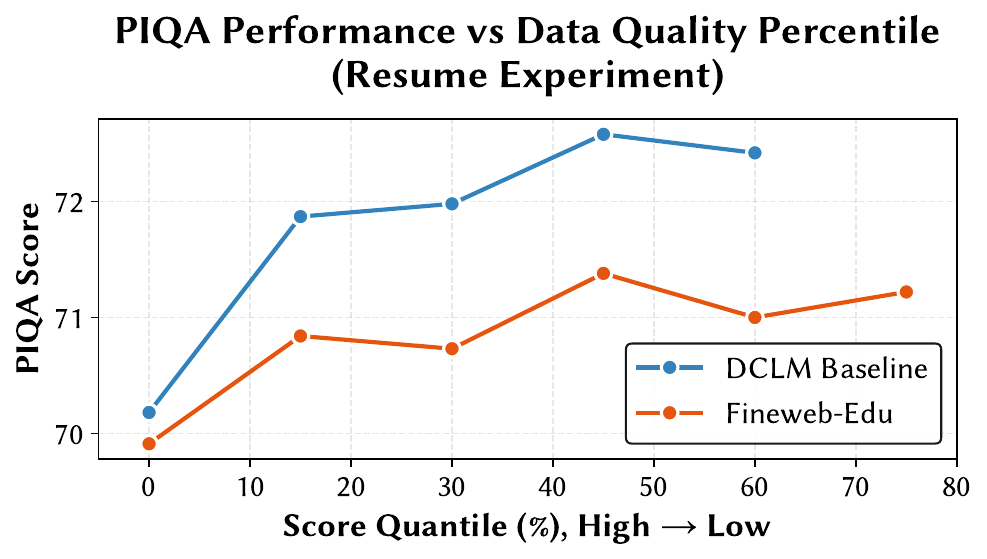} 
    \end{subfigure}
    \begin{subfigure}[b]{0.49\linewidth}
    \includegraphics[width=\linewidth]{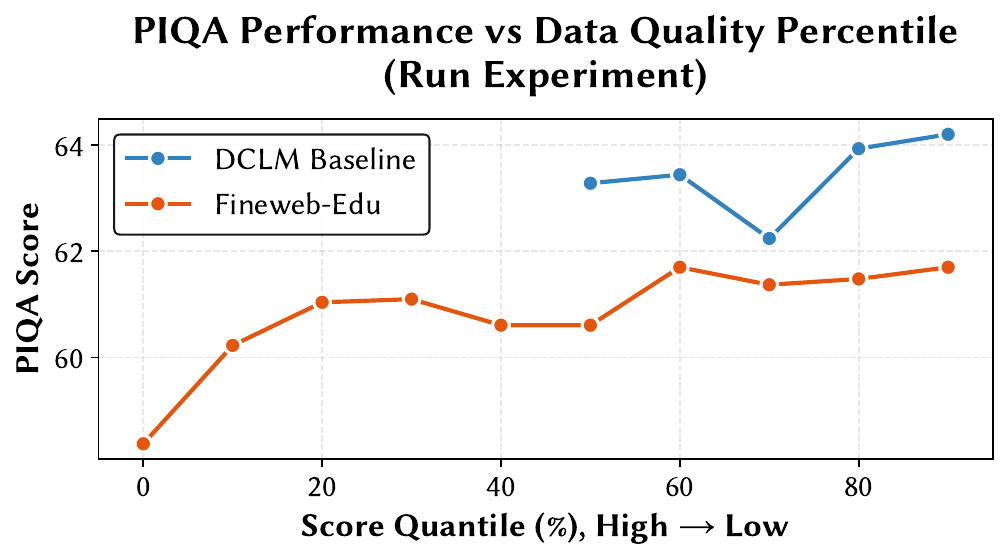} 
    \end{subfigure}
    \\
    \begin{subfigure}[b]{0.49\linewidth}
    \includegraphics[width=\linewidth]{benchmark_plots_resume/WinoGrande_resume.pdf} 
    \end{subfigure}
    \begin{subfigure}[b]{0.49\linewidth}
    \includegraphics[width=\linewidth]{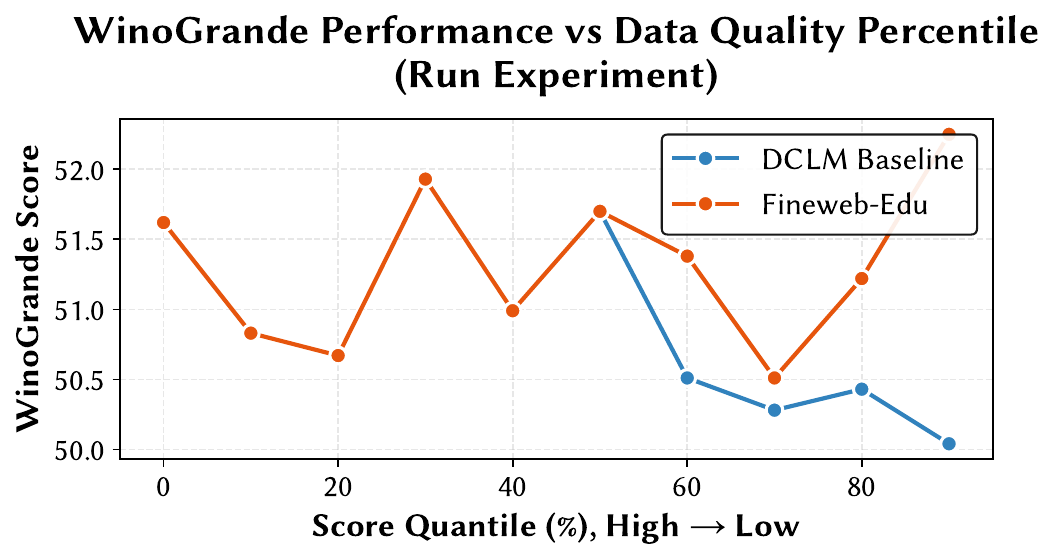} 
    \end{subfigure}
    \caption{Quantile Benchmarks: DCLM-Baseline is better on understanding-oriented benchmarks.}
    \label{fig:quantile_benchmark_understanding}
\end{figure}

\begin{figure}[htbp]
    \centering
    \begin{subfigure}[b]{0.49\linewidth}
    \includegraphics[width=\linewidth]{benchmark_plots_resume/MMLU_resume.pdf} 
    \end{subfigure}
    \begin{subfigure}[b]{0.49\linewidth}
    \includegraphics[width=\linewidth]{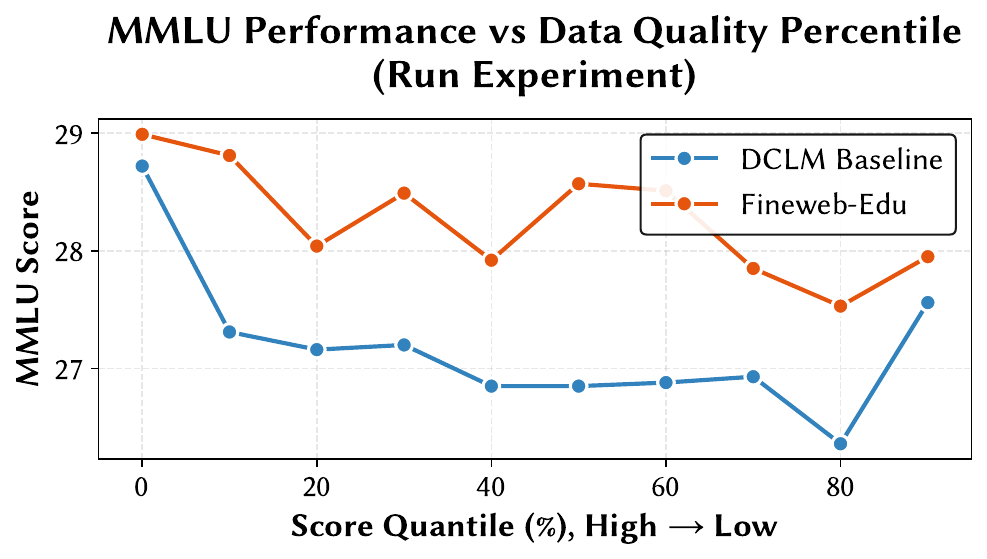} 
    \end{subfigure}
    \\
    \begin{subfigure}[b]{0.49\linewidth}
    \includegraphics[width=\linewidth]{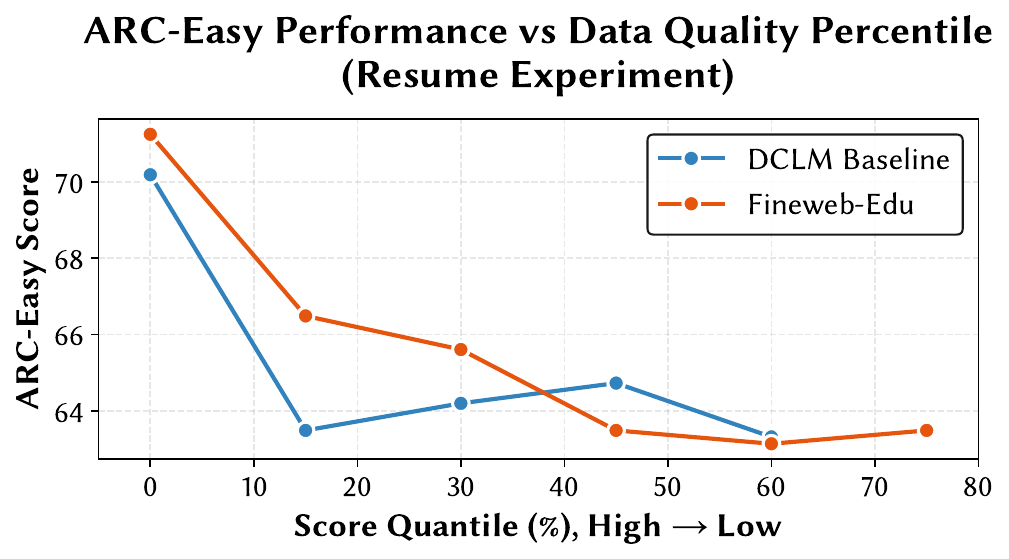} 
    \end{subfigure}
    \begin{subfigure}[b]{0.49\linewidth}
    \includegraphics[width=\linewidth]{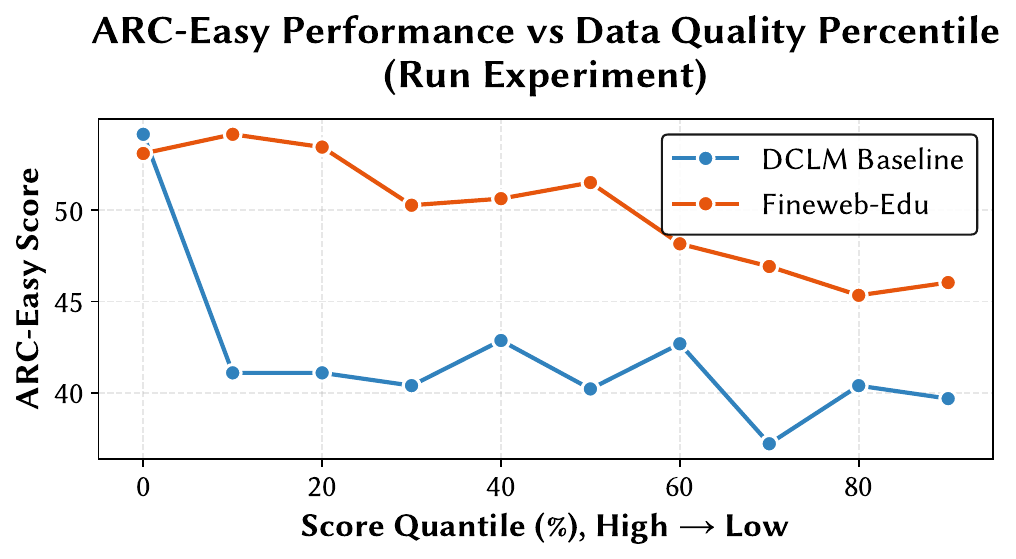} 
    \end{subfigure}
    \\
    \begin{subfigure}[b]{0.49\linewidth}
    \includegraphics[width=\linewidth]{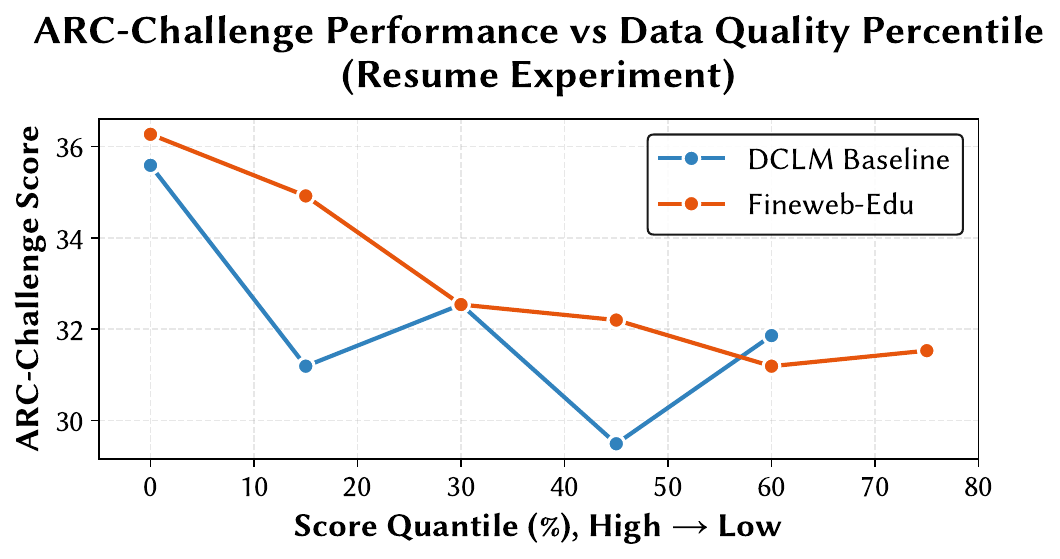} 
    \end{subfigure}
    \begin{subfigure}[b]{0.49\linewidth}
    \includegraphics[width=\linewidth]{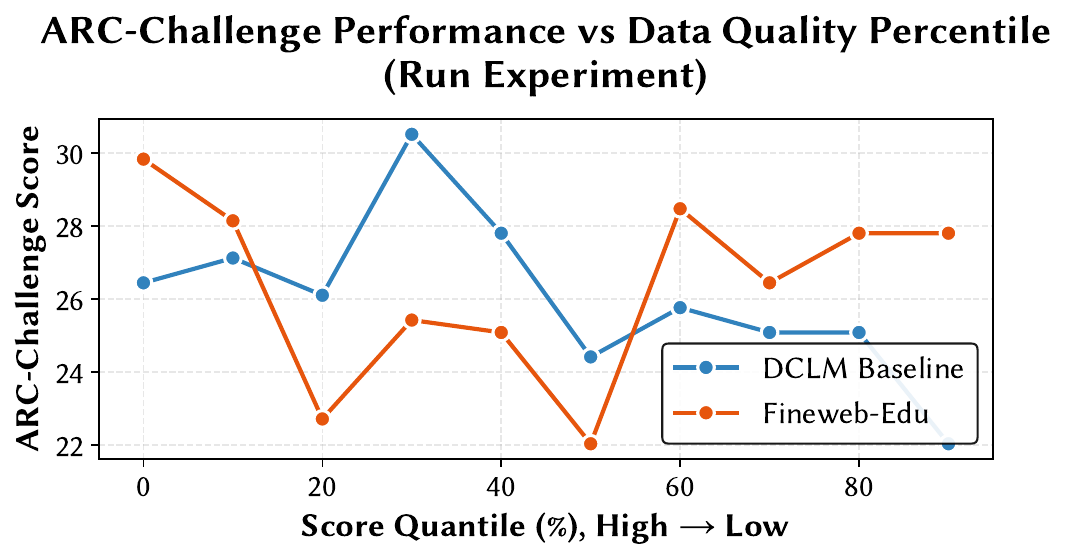} 
    \end{subfigure}    
    \\
    \begin{subfigure}[b]{0.49\linewidth}
    \includegraphics[width=\linewidth]{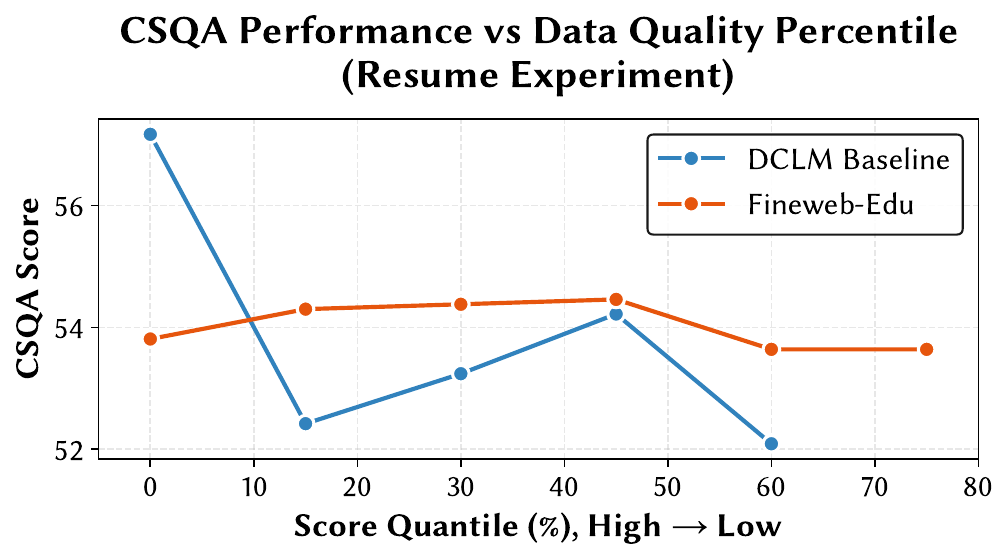} 
    \end{subfigure}
    \begin{subfigure}[b]{0.49\linewidth}
    \includegraphics[width=\linewidth]{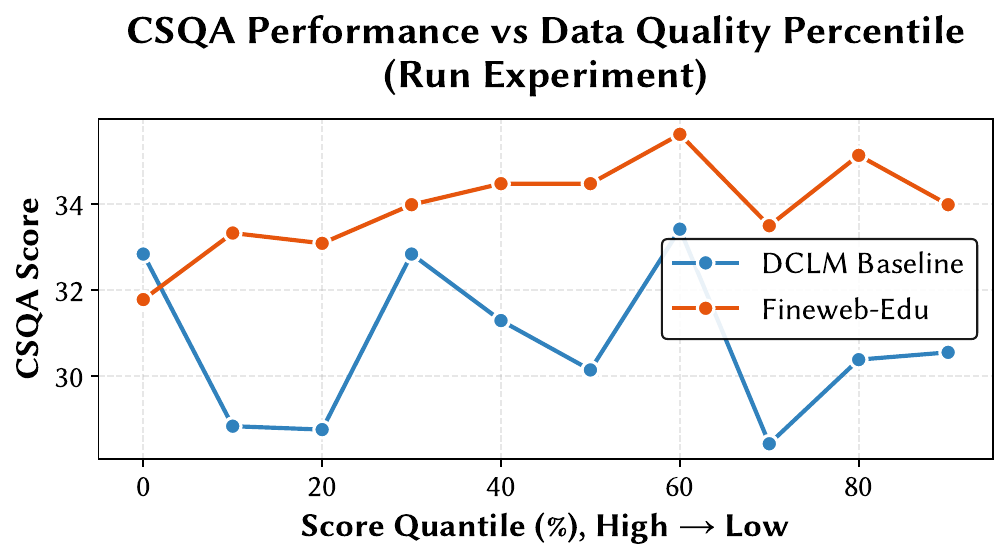} 
    \end{subfigure}
    \\
    \begin{subfigure}[b]{0.49\linewidth}
    \includegraphics[width=\linewidth]{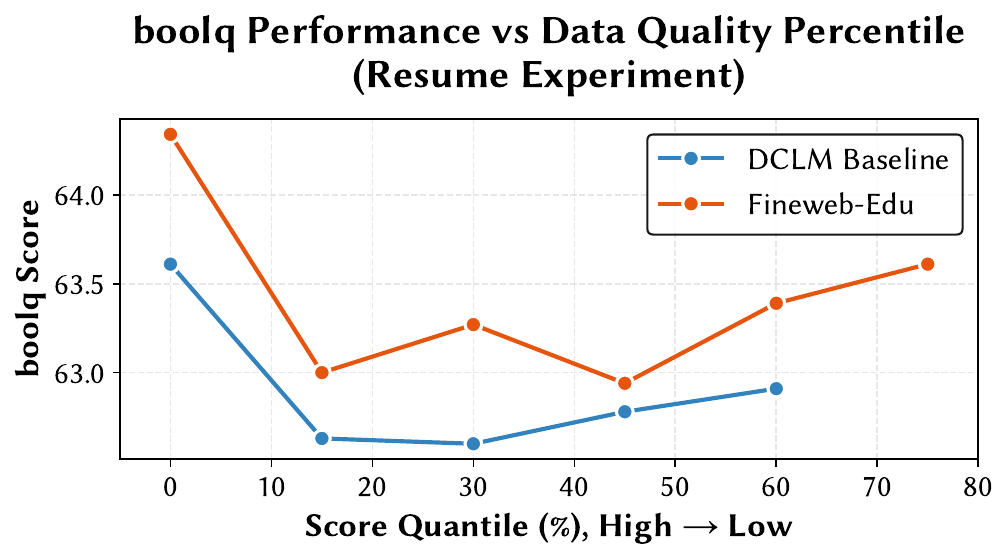} 
    \end{subfigure}
    \begin{subfigure}[b]{0.49\linewidth}
    \includegraphics[width=\linewidth]{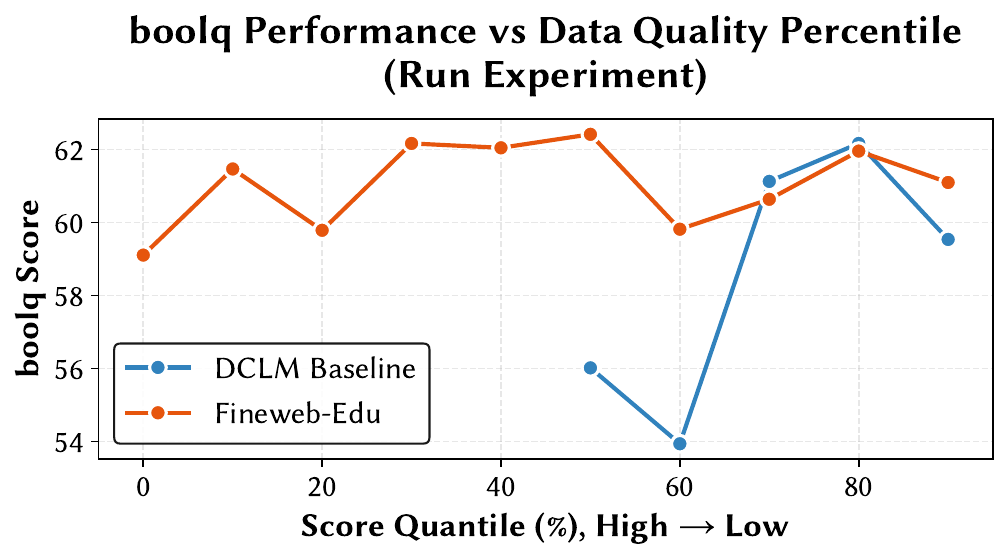} 
    \end{subfigure}
    \caption{Quantile Benchmarks: FineWeb-Edu is better on knowledge-oriented benchmarks.}
    \label{fig:quantile_benchmarks_knowledge}
\end{figure}

We show full quantile benchmarking results in \Cref{fig:quantile_benchmark_understanding,fig:quantile_benchmarks_knowledge}. The overall observations are discussed in \Cref{sec:data_benchmark} in detail. The DCLM-Baseline leading experiments are shown in \Cref{fig:quantile_benchmark_understanding} and Fineweb-Edu leading experiments are shown in \Cref{fig:quantile_benchmarks_knowledge}. 

\FloatBarrier

\section{Datasets Used in Training}\label{sec:dataset_used}

\Cref{tab:datasets} is a comprehensive list of all datasets used in the training process of \fullsys. All datasets are publicly available to acquire, and most of them are hosted on Hugging Face unless otherwise noted.

\let\OldTblrNote\TblrNote
\renewcommand{\TblrNote}[1]{\textcolor{teal}{\OldTblrNote{#1}}}

\begin{longtblr}[
  caption = {All Datasets Used in the Training of \fullsys.},
  label = {tab:datasets},
  note{0} = {Token counts are pre-deduplication rough numbers. They may differ from the well-known ones due to partial inclusion of mixed datasets, the use of different revisions/splits/tokenizers, or some other pre-processing.},
  note{1} = {This dataset originates from Common Crawl and thereby abides by its terms of use~\cite{commoncrawltou}.},
  note{2} = {This dataset contains source code with various licenses.},
  note{3} = {The license has changed over time, according to \url{https://stackoverflow.com/help/licensing}.},
  note{4} = {This dataset is created by mixing and de-duplicating all source datasets.},
  note{5} = {This dataset is acquired from OpenDataLab (\url{https://opendatalab.com}).},
  note{6} = {Some old content of Wikipedia is dual-licensed under CC BY 4.0 and GFDL.},
]{
  colspec = {X[0.77,l] c X[2.1,l] c X[1.9,c]}, 
  width = \linewidth,
  rowhead = 1,
  rows = {valign=m}, %
  row{1} = {font=\bfseries, bg=gray!10}, %
  hlines, %
}

Name & Type & Hugging Face ID & \#Tokens\TblrNote{0} & License(s) \\

DCLM-Baseline & English & mlfoundations/dclm-baseline-1.0~\cite{li2024datacomplm} & 4T & CC BY 4.0\TblrNote{1} \\
FineWiki-EN & English & HuggingFaceFW/finewiki~\cite{penedo2025finewiki} & 8.7B & CC BY-SA 4.0\TblrNote{6} \\
FinePDFs & English & HuggingFaceFW/finepdfs~\cite{kydlicek2025finepdfs} & 3T & ODC-By 1.0\TblrNote{1} \\
Flan & English & allenai/dolmino-mix-1124 & 17B & ODC-By 1.0 \\
Pes2O & English & allenai/dolmino-mix-1124 & 58.6B & ODC-By 1.0 \\
FineWeb-Edu-EN & English & HuggingFaceTB/smollm-corpus~\cite{benallal2024smollmcorpus} & 220B & ODC-By 1.0\TblrNote{1} \\
ArXiv & English & togethercomputer/RedPajama-Data-1T~\cite{together2023redpajama} & 28B & {Metadata: CC0 1.0~\cite{arxivTermsArXiv} \\ Content: various~\cite{arxivLicenseCopyright}} \\
Cosmopedia-v2 & English & HuggingFaceTB/smollm-corpus~\cite{benallal2024smollmcorpus} & 27B & ODC-By 1.0 \\

FineWiki-CN & Chinese & HuggingFaceFW/finewiki~\cite{penedo2025finewiki} & 1.1B & CC BY-SA 4.0\TblrNote{6} \\
Fineweb-Edu-CN & Chinese & opencsg/Fineweb-Edu-Chinese-V2.1~\cite{yu2025opencsgchinesecorpusseries} & 1.5T & OpenCSG Community License~\cite{opencsglicense}, Apache 2.0 \\
Baidu-Baike & Chinese & mohamedah/baidu\_baike & 1.2B & MIT \\
UNDL ZH-EN Aligned & Chinese & bot-yaya/undl\_zh2en\_aligned & 1.8B & MIT \\
Dedup-Merged-PAC-CN\TblrNote{4} & Chinese & {BAAI/CCI-Data \\ BAAI/CCI2-Data \\ BAAI/CCI3-Data~\cite{wang2024cci30hqlargescalechinesedataset} \\ Skywork/SkyPile-150B~\cite{wei2023skywork} \\ OpenDataLab/WanJuan1.0~\cite{he2023wanjuan, he2024opendatalabempoweringgeneralartificial}\TblrNote{5} \\ BAAI/IndustryCorpus \\ BAAI/IndustryCorpus2~\cite{industrycorpus2} \\ WuDaoCorpus2.0~\cite{ZHANG2021216, ZHANG202193}\TblrNote{5}} & 178B & {CCI\{,2,3\}-Data: CCI Usage Agreement~\cite{cciagreement} \\ SkyPile-150B: Skywork Community License~\cite{skyworklicense}, Apache 2.0 \\ WanJuan1.0: CC BY-4.0 \\ IndustryCorpus\{,2\}: Apache 2.0 \\ WuDaoCorpus2.0: Apache 2.0} \\

OpenWebMath & Math & open-web-math/open-web-math~\cite{paster2023openwebmath} & 14.7B & ODC-By 1.0\TblrNote{1} \\
FineMath & Math & HuggingFaceTB/finemath~\cite{allal2025smollm2smolgoesbig} & 10B & ODC-By 1.0 \\
MegaMath-Web-Pro & Math & LLM360/MegaMath~\cite{zhou2025megamath} & 300B & ODC-By 1.0 \\
AutoMathText & Math & math-ai/AutoMathText~\cite{zhang-etal-2025-autonomous} & 8.7B & CC BY-SA 4.0 \\
SwallowMath-v2 & Math & tokyotech-llm/swallow-math-v2~\cite{fujii2025rewritingpretrainingdataboosts} & 32B & Apache 2.0 \\

StarCoder & Code & bigcode/starcoderdata~\cite{Kocetkov2022TheStack} & 250B & Original Licenses\TblrNote{2} \\
Stack V2 Smol & Code & bigcode/the-stack-v2~\cite{lozhkov2024starcoder} & 900B & Original Licenses\TblrNote{2} \\
StackExchange & Code & togethercomputer/RedPajama-Data-1T~\cite{together2023redpajama} & 20B & CC BY-SA 2.5/3.0/4.0\TblrNote{3} \\
Python-Edu & Code & HuggingFaceTB/smollm-corpus~\cite{lozhkov2024starcoder, benallal2024smollmcorpus} & 3.4B & ODC-By 1.0, Original Licenses\TblrNote{2} \\
Algebraic-Stack & Code & typeof/algebraic-stack~\cite{azerbayev2023llemma, paster2023openwebmath} & 11B & ODC-By 1.0\TblrNote{1} \\
Swallow-Code-v2 & Code & tokyotech-llm/swallow-code-v2~\cite{fujii2025rewritingpretrainingdataboosts} & 49.8B & Apache 2.0 \\

SlimOrca & SFT & Open-Orca/SlimOrca~\cite{mukherjee2023orca,longpre2023flan} & 190M & MIT \\
JiuZhang3.0-Corpus-CoT & SFT & ToheartZhang/JiuZhang3.0-Corpus-CoT~\cite{zhou2024jiuzhang30} & 358B & \emph{Not Specified} \\
Tulu-3-Sft-0225 & SFT & allenai/tulu-3-sft-mixture~\cite{lambert2024tulu3} & 640M & ODC-By 1.0 (mixed) \\
downstream\TblrNote{4} & SFT & {%
cais/mmlu~\cite{mmlu,hendrycks2021ethics} \\
openai/gsm8k~\cite{gsm8k} \\
allenai/ai2\_arc~\cite{arc} \\
allenai/openbookqa~\cite{openbookqa} \\
Rowan/hellaswag~\cite{hellaswag} \\
allenai/winogrande~\cite{winogrande}%
} & 12.6M & {%
MMLU, GSM8K: MIT \\
ai2\_arc: CC BY-SA 4.0 \\
OpenBookQA: \emph{Not Specified} \\
hellaswag: MIT \\
winogrande: \emph{Not Specified}%
} \\

\end{longtblr}

\let\TblrNote\OldTblrNote

To enhance the reproducibility of our results and accessibility, we have conducted careful screening and selection of datasets at the best of our ability. We would like to ensure that our model (\sys) and training datasets are compliant with all licenses and agreements presented in \Cref{tab:datasets}, so that they can be released under a permissive license for the community to use (still on an ``as-is'' and ``use-at-your-own-risk'' basis). Everyone can use these same datasets to reproduce our results and further adapt and/or publish both the modified datasets and models at will, free from potential legal risk.

For example, although the Nemotron series datasets from NVIDIA are also available on Hugging Face upon request, the \emph{NVIDIA Data Agreement for Model Training}~\cite{nvidiamodellicense} applied to them disallows redistribution, and even public display of the dataset. Therefore, they are fully excluded from our training data.

\section{Phase-wise Data Mixture}
\label{sec:data_mixture_details}

\begin{figure}[htbp]
    \centering
    \includegraphics[width=\linewidth]{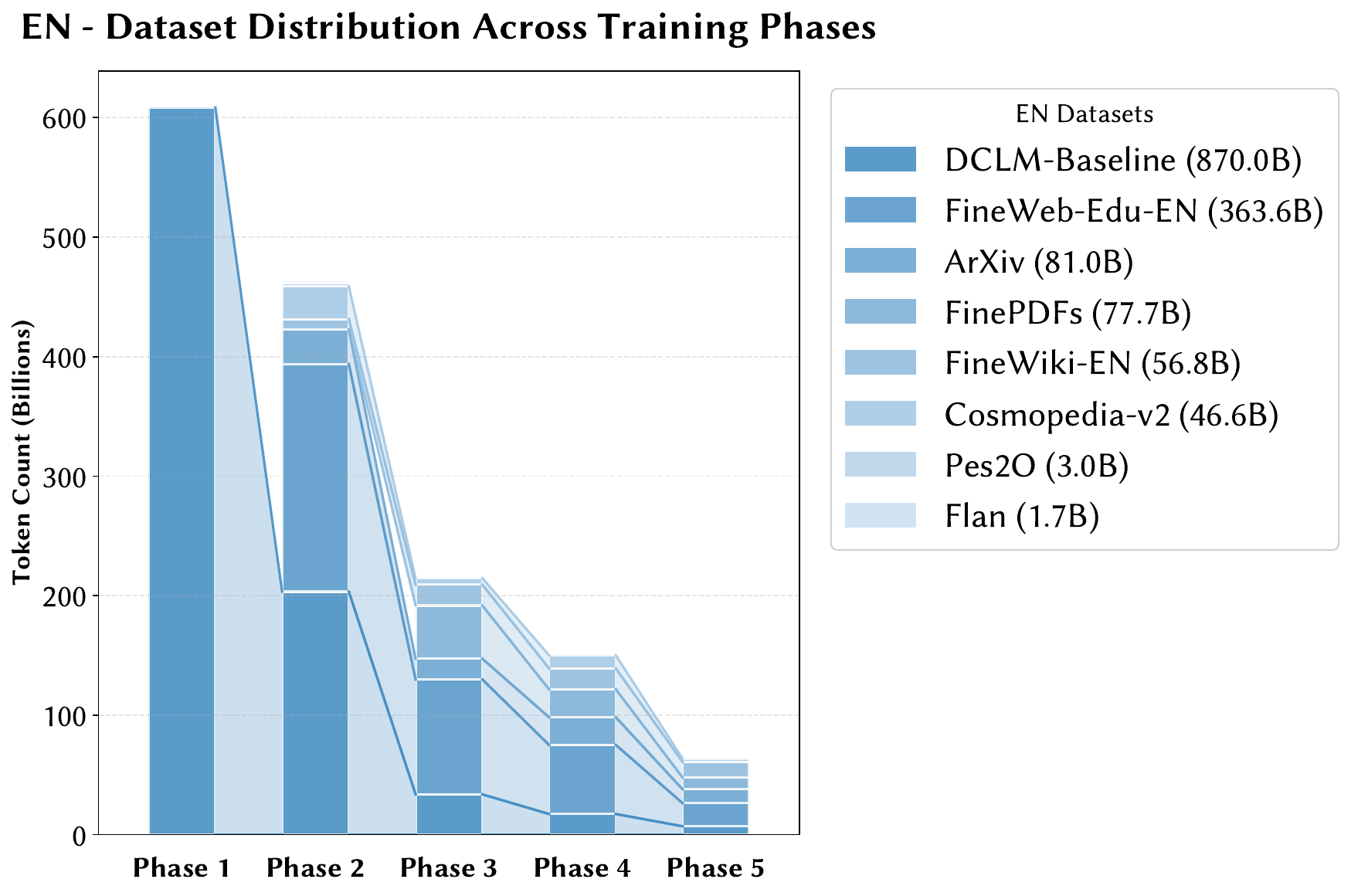}
    \caption{Phase-wise Dataset Mixture: English.}
    \label{fig:en_mixture_pipeline}
\end{figure}

\begin{figure}[htbp]
    \centering
    \includegraphics[width=\linewidth]{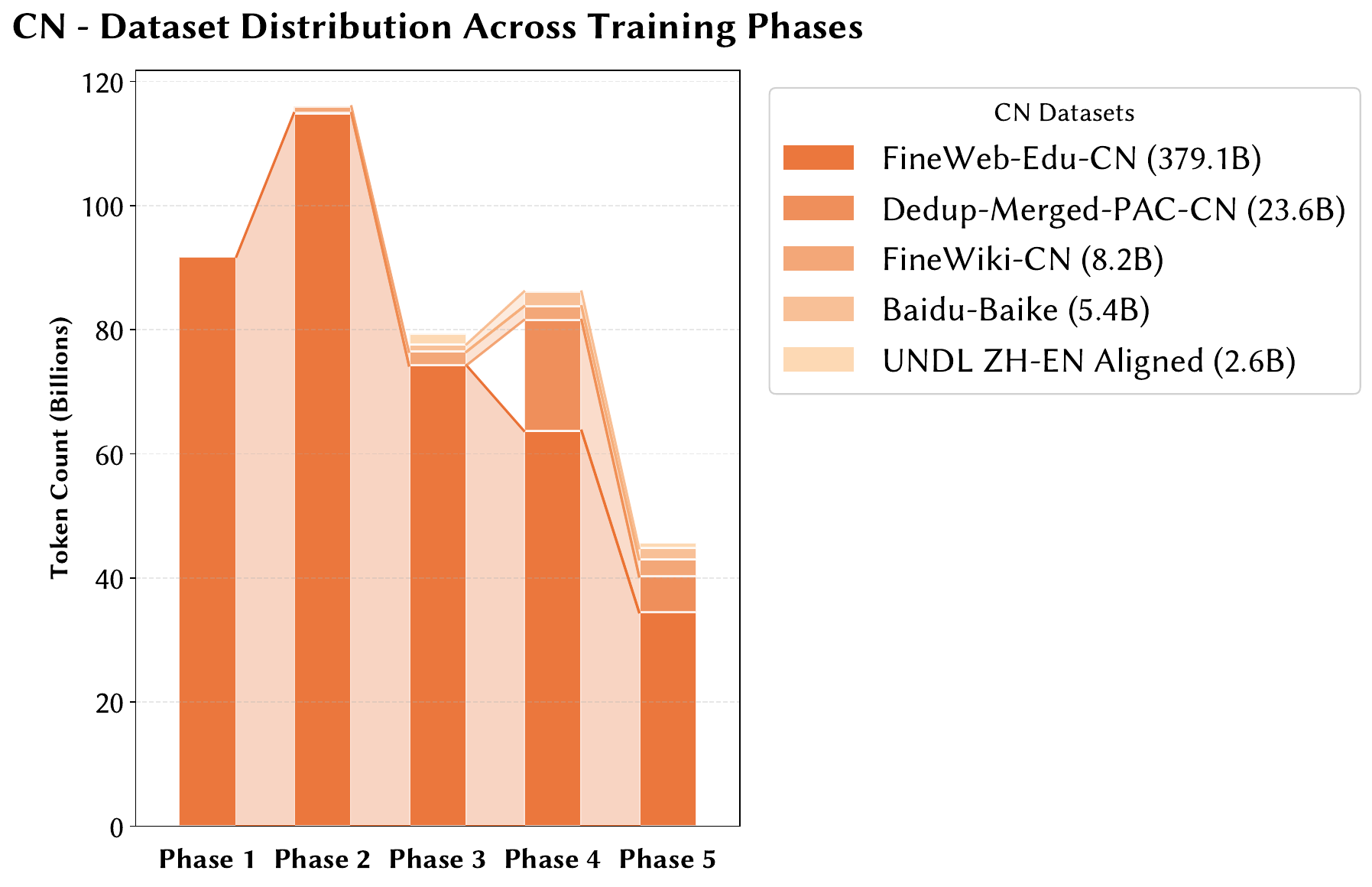}
    \caption{Phase-wise Dataset Mixture: Chinese.}
    \label{fig:cn_mixture_pipeline}
\end{figure}

\begin{figure}[htbp]
    \centering
    \includegraphics[width=\linewidth]{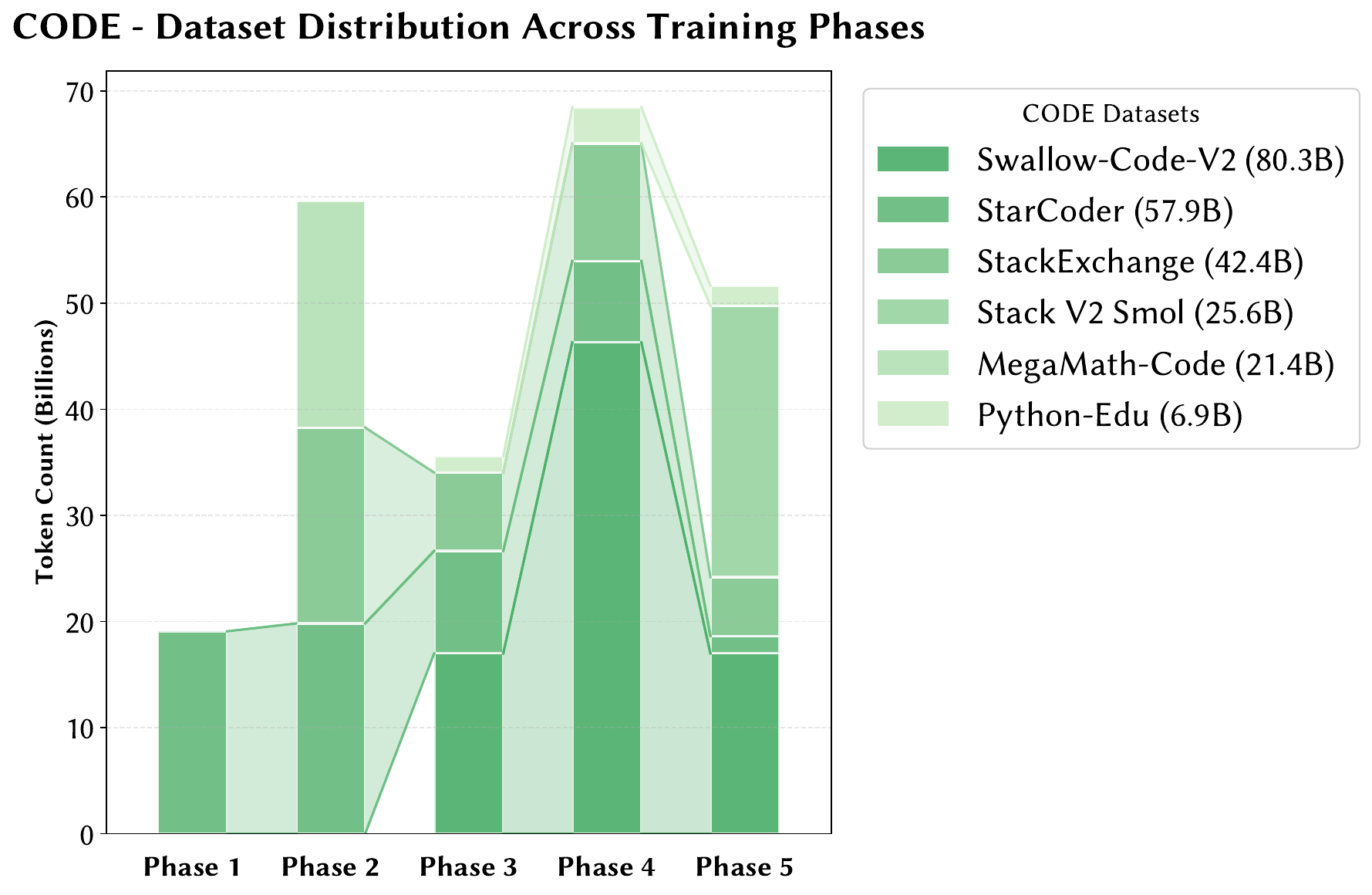}
    \caption{Phase-wise Dataset Mixture: Code.}
    \label{fig:code_mixture_pipeline}
\end{figure}

\begin{figure}[htbp]
    \centering
    \includegraphics[width=\linewidth]{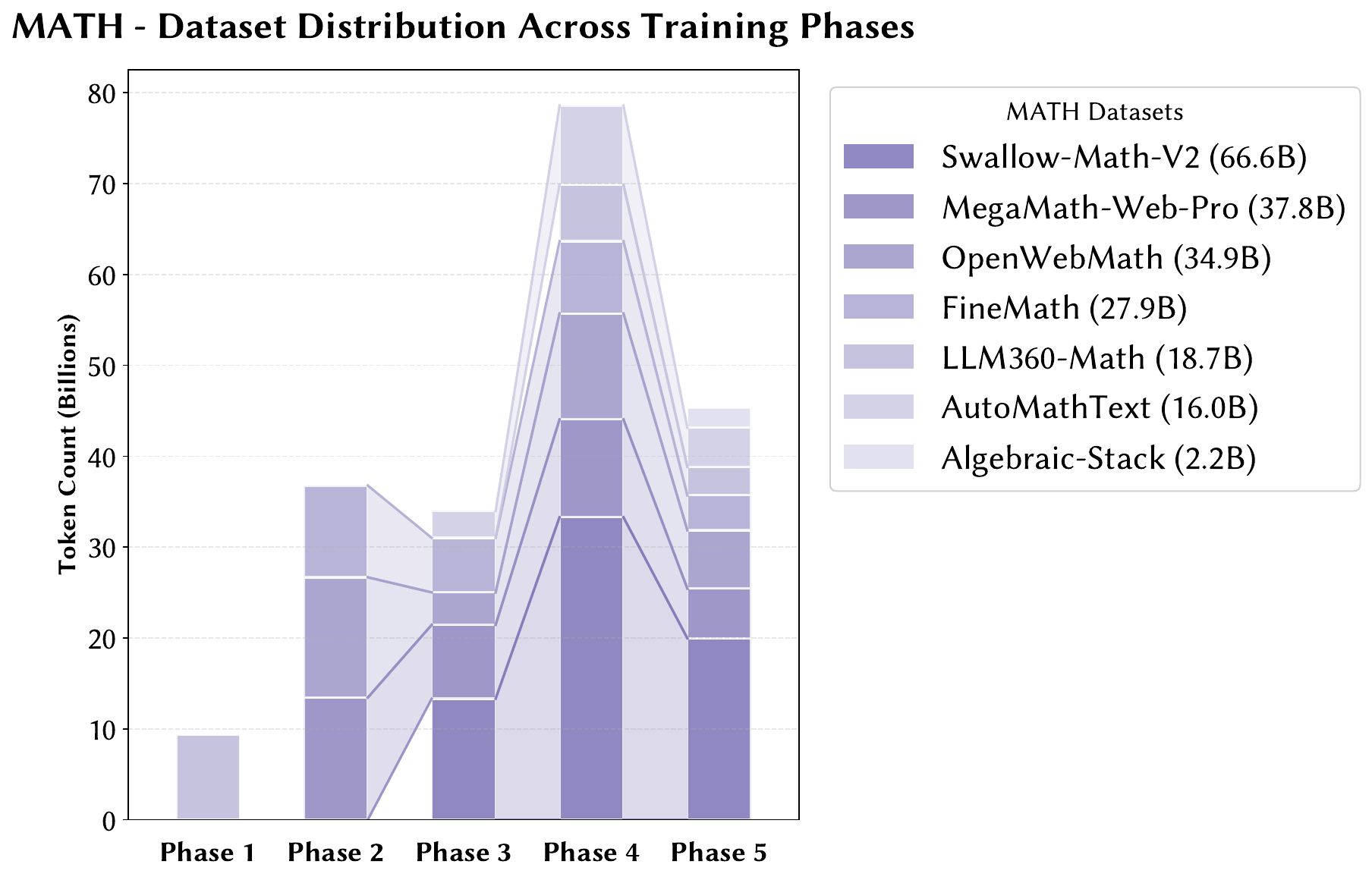}
    \caption{Phase-wise Dataset Mixture: Math.}
    \label{fig:math_mixture_pipeline}
\end{figure}

\begin{figure}[htbp]
    \centering
    \includegraphics[width=\linewidth]{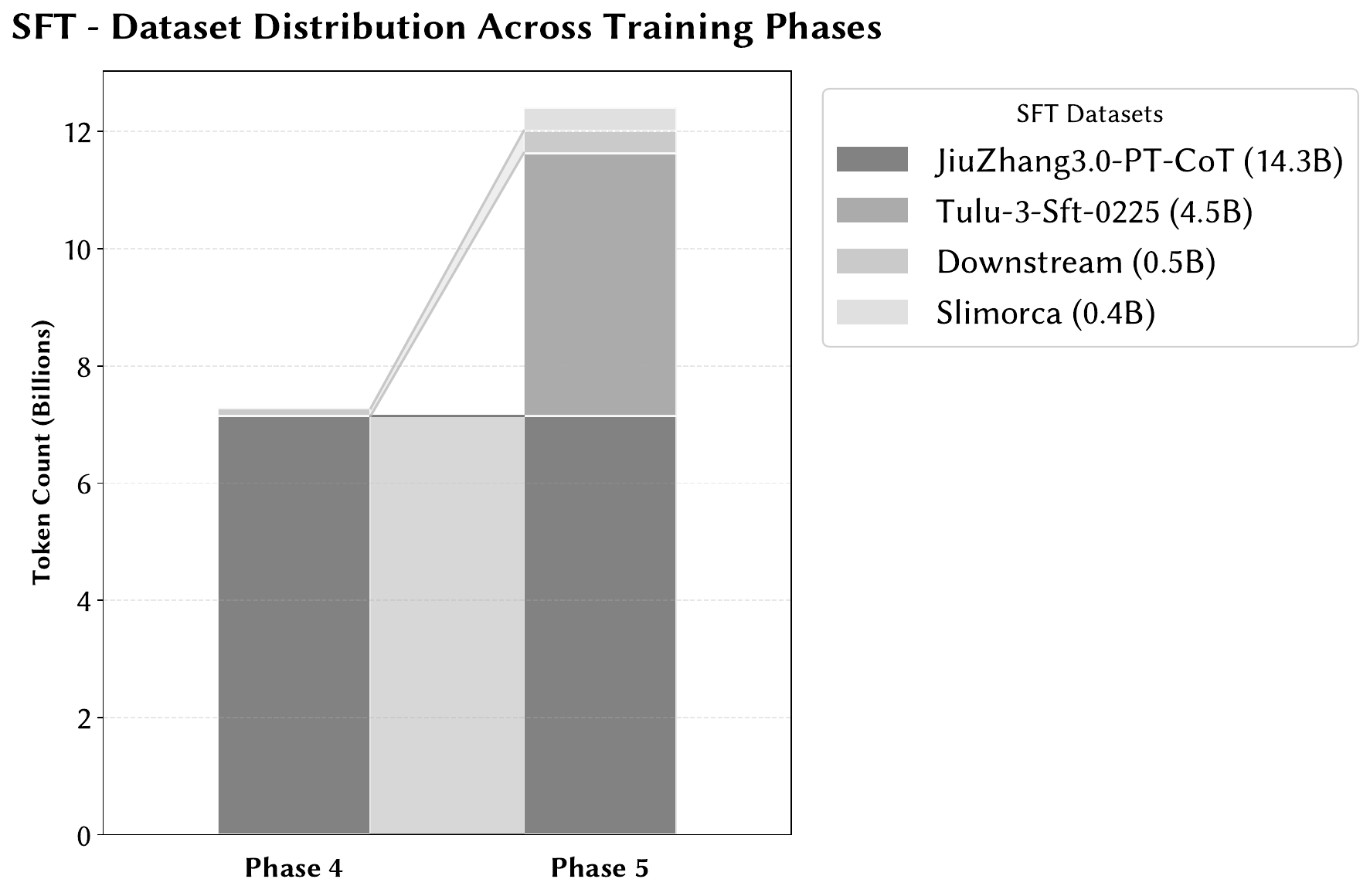}
    \caption{Phase-wise Dataset Mixture: SFT.}
    \label{fig:sft_mixture_pipeline}
\end{figure}

In this section, we first visualize the dataset counts within each domain throughout multi-phase training. The transitions of the English, Chinese, Math, Code, and SFT datasets are shown in \Cref{fig:en_mixture_pipeline,fig:cn_mixture_pipeline,fig:math_mixture_pipeline,fig:code_mixture_pipeline,fig:sft_mixture_pipeline}, respectively. Moreover, we list the detailed dataset composition for each phase in \Cref{tab:phase1_stats,tab:phase2_stats,tab:phase3_stats,tab:phase4_stats,tab:phase5_stats}, from Phase~1 to Phase~5. In these tables, there are four primary cases:
\begin{enumerate}
    \item The entire dataset is used in this phase. The score column is denoted as \textit{(fully used)}, and the actual ratio is $100.0\%$, such as DCLM-Baseline in Phase~1 (\Cref{tab:phase1_stats}) and Fineweb-Edu-EN in Phase~2 (\Cref{tab:phase2_stats}).
    \item The dataset is filtered according to its specific score column (\textit{Score Col} in the tables), retaining only top-scoring samples with an \textit{Actual Ratio}. For example, Fineweb-Edu-CN in Phase~1 keeps the top $20.8\%$ of \textit{score} (\Cref{tab:phase1_stats}), and StarCoder in Phase~2 keeps the top $10.4\%$ of \textit{max\_stars\_count}.
    \item The dataset has no quality metrics, and we randomly select samples accounting for the \textit{Actual Ratio}. For example, we randomly select $10.0\%$ of samples from StarCoder and $30.0\%$ from LLM360-Math in Phase~1 (\Cref{tab:phase1_stats}).
    \item The dataset is repeated within the phase. The score column is denoted as \textit{duplicate}, and the actual ratio exceeds $100\%$. The repetition count is determined by rounding the actual ratio according to its decimal part. For example, FineWiki-CN is repeated twice in Phase~3 (\Cref{tab:phase3_stats}), and for Baidu-Baike in Phase~5, we round $1.5$ to either 1 or 2 with equal probability, then repeat the samples that many times (\Cref{tab:phase5_stats}).
\end{enumerate}

In addition, LLM360-Math is a deduplicated subset of the MegaMath dataset~\citep{zhou2025megamath}, and we select only the top $5\%$ of rows from the English partition of the FinePDFs dataset~\citep{kydlicek2025finepdfs}, according to Fineweb-Edu classifier scores~\citep{fineweb}.


\begin{table}[htbp]
\centering
\caption{Phase 1 Dataset Statistics.}
\label{tab:phase1_stats}
\begin{narrowtblr}{l c c c c}
Dataset & Score Col & Token Before (B) & Token After (B) & Actual Ratio \\
DCLM-Baseline & (fully used) & 608.54 & 608.54 & $100.0\%$ \\
FineWeb-Edu-CN & score & 441.66 & 91.78 & $20.8\%$ \\
StarCoder & random & 190.60 & 19.08 & $10.0\%$ \\
LLM360-Math & random & 31.12 & 9.34 & $30.0\%$ \\
\end{narrowtblr}
\end{table}

\begin{table}[htbp]
\centering
\caption{Phase 2 Dataset Statistics.}
\label{tab:phase2_stats}
\begin{narrowtblr}{l c c c c}
Dataset & Score Col & Token Before (B) & Token After (B) & Actual Ratio \\
FineWeb-Edu-CN & score & 441.66 & 114.88 & $26.0\%$ \\
FineWiki-CN & (fully used) & 1.10 & 1.10 & $100.0\%$ \\
FineWeb-Edu-EN & (fully used) & 190.37 & 190.37 & $100.0\%$ \\
DCLM-Baseline & fasttext score & 608.54 & 203.32 & $33.4\%$ \\
Flan & random & 17.15 & 1.71 & $10.0\%$ \\
Pes2O & random & 60.11 & 3.00 & $5.0\%$ \\
FineWiki-EN & (fully used) & 8.74 & 8.74 & $100.0\%$ \\
ArXiv & (fully used) & 28.93 & 28.93 & $100.0\%$ \\
Cosmopedia-v2 & (fully used) & 27.41 & 27.41 & $100.0\%$ \\
FineMath & (fully used) & 10.10 & 10.10 & $100.0\%$ \\
OpenWebMath & (fully used) & 13.23 & 13.23 & $100.0\%$ \\
MegaMath-Web-Pro & (fully used) & 13.45 & 13.45 & $100.0\%$ \\
StackExchange & (fully used) & 18.46 & 18.46 & $100.0\%$ \\
MegaMath-Code & random & 42.77 & 21.38 & $50.0\%$ \\
StarCoder & max\_stars\_count & 190.60 & 19.82 & $10.4\%$ \\
\end{narrowtblr}
\end{table}

\begin{table}[htbp]
\centering
\caption{Phase 3 Dataset Statistics.}
\label{tab:phase3_stats}
\begin{narrowtblr}{l c c c c}
Dataset & Score Col & Token Before (B) & Token After (B) & Actual Ratio \\
FineWeb-Edu-CN & score & 441.66 & 74.26 & $16.8\%$ \\
FineWiki-CN & duplicate & 1.10 & 2.20 & $200.0\%$ \\
UNDL ZH-EN Aligned & (fully used) & 1.75 & 1.75 & $100.0\%$ \\
Baidu-Baike & (fully used) & 1.19 & 1.19 & $100.0\%$ \\
FineWeb-Edu-EN & score & 190.37 & 96.13 & $50.5\%$ \\
DCLM-Baseline & fasttext score & 608.54 & 33.77 & $5.5\%$ \\
FineWiki-EN & duplicate & 8.74 & 17.47 & $200.0\%$ \\
ArXiv & random & 28.93 & 17.35 & $60.0\%$ \\
FineMath & score & 10.10 & 6.00 & $59.4\%$ \\
MegaMath-Web-Pro & math\_score & 13.45 & 8.13 & $60.4\%$ \\
StackExchange & random & 18.46 & 7.38 & $40.0\%$ \\
StarCoder & max\_stars\_count & 190.60 & 9.65 & $5.1\%$ \\
Swallow-Code-V2 & score & 50.62 & 17.00 & $33.6\%$ \\
Python-Edu & score & 3.41 & 1.56 & $45.7\%$ \\
Cosmopedia-v2 & random & 27.41 & 5.48 & $20.0\%$ \\
AutoMathText & lm\_q1q2\_score & 8.71 & 2.97 & $34.1\%$ \\
OpenWebMath & math\_score & 13.23 & 3.57 & $27.0\%$ \\
Swallow-Math-V2 & random & 33.29 & 13.32 & $40.0\%$ \\
FinePDFs & (fully used) & 44.50 & 44.50 & $100.0\%$ \\
\end{narrowtblr}
\end{table}

\begin{table}[htbp]
\centering
\caption{Phase 4 Dataset Statistics.}
\label{tab:phase4_stats}
\begin{narrowtblr}{l c c c c}
Dataset & Score Col & Token Before (B) & Token After (B) & Actual Ratio \\
FineWeb-Edu-CN & score & 441.66 & 63.71 & $14.4\%$ \\
FineWiki-CN & duplicate & 1.10 & 2.20 & $200.0\%$ \\
Baidu-Baike & duplicate & 1.19 & 2.39 & $200.0\%$ \\
FineWeb-Edu-EN & score & 190.37 & 57.79 & $30.4\%$ \\
DCLM-Baseline & fasttext score & 608.54 & 17.32 & $2.8\%$ \\
FineWiki-EN & duplicate & 8.74 & 17.47 & $200.0\%$ \\
ArXiv & random & 28.93 & 23.15 & $80.0\%$ \\
FineMath & score & 10.10 & 7.95 & $78.7\%$ \\
MegaMath-Web-Pro & math\_score & 13.45 & 10.76 & $80.0\%$ \\
StackExchange & random & 18.46 & 11.07 & $60.0\%$ \\
StarCoder & max\_stars\_count & 190.60 & 7.67 & $4.0\%$ \\
Downstream & duplicate & 0.01 & 0.13 & $1000.0\%$ \\
Swallow-Code-V2 & score & 50.62 & 46.30 & $91.5\%$ \\
Python-Edu & (fully used) & 3.41 & 3.41 & $100.0\%$ \\
Cosmopedia-v2 & random & 27.41 & 10.97 & $40.0\%$ \\
AutoMathText & (fully used) & 8.71 & 8.71 & $100.0\%$ \\
LLM360-Math & random & 31.12 & 6.22 & $20.0\%$ \\
OpenWebMath & math\_score & 13.23 & 11.66 & $88.1\%$ \\
Swallow-Math-V2 & (fully used) & 33.29 & 33.29 & $100.0\%$ \\
JiuZhang3.0-PT-CoT & duplicate & 3.58 & 7.15 & $200.0\%$ \\
FinePDFs & fineweb-edu-classifier & 44.50 & 23.38 & $52.5\%$ \\
Dedup-Merged-PAC-CN & random & 178.49 & 17.85 & $10.0\%$ \\
\end{narrowtblr}
\end{table}

\begin{table}[htbp]
\centering
\caption{Phase 5 Dataset Statistics.}
\label{tab:phase5_stats}
\begin{narrowtblr}{l c c c c}
Dataset & Score Col & Token Before (B) & Token After (B) & Actual Ratio \\
FineWeb-Edu-CN & score & 441.66 & 34.50 & $7.8\%$ \\
FineWiki-CN & duplicate & 1.10 & 2.75 & $250.0\%$ \\
UNDL ZH-EN Aligned & random & 1.75 & 0.88 & $50.3\%$ \\
Baidu-Baike & duplicate & 1.19 & 1.79 & $150.0\%$ \\
FineWeb-Edu-EN & score & 190.37 & 19.35 & $10.2\%$ \\
DCLM-Baseline & fasttext score & 608.54 & 7.06 & $1.2\%$ \\
FineWiki-EN & duplicate & 8.74 & 13.10 & $150.0\%$ \\
ArXiv & random & 28.93 & 11.60 & $40.1\%$ \\
FineMath & score & 10.10 & 3.86 & $38.2\%$ \\
MegaMath-Web-Pro & math\_score & 13.45 & 5.47 & $40.7\%$ \\
StackExchange & random & 18.46 & 5.54 & $30.0\%$ \\
StarCoder & max\_stars\_count & 190.60 & 1.64 & $0.9\%$ \\
Downstream & duplicate & 0.01 & 0.38 & $3000.0\%$ \\
Swallow-Code-V2 & score & 50.62 & 17.00 & $33.6\%$ \\
Python-Edu & score & 3.41 & 1.92 & $56.3\%$ \\
Cosmopedia-v2 & random & 27.41 & 2.74 & $10.0\%$ \\
AutoMathText & lm\_q1q2\_score & 8.71 & 4.32 & $49.6\%$ \\
LLM360-Math & random & 31.12 & 3.11 & $10.0\%$ \\
OpenWebMath & math\_score & 13.23 & 6.41 & $48.5\%$ \\
Swallow-Math-V2 & random & 33.29 & 19.97 & $60.0\%$ \\
JiuZhang3.0-PT-CoT & duplicate & 3.58 & 7.15 & $200.0\%$ \\
FinePDFs & fineweb-edu-classifier & 44.50 & 9.86 & $22.2\%$ \\
Dedup-Merged-PAC-CN & pac\_score & 178.49 & 5.77 & $3.2\%$ \\
Tulu-3-Sft-0225 & duplicate & 0.64 & 4.48 & $700.0\%$ \\
Stack V2 Smol & random & 127.98 & 25.56 & $20.0\%$ \\
Slimorca & duplicate & 0.20 & 0.40 & $200.0\%$ \\
Algebraic-Stack & max\_stars\_count & 8.51 & 2.17 & $25.5\%$ \\
\end{narrowtblr}
\end{table}

\FloatBarrier

\section{Experimental Settings}
\label{sec:experiment_setting_details}

\subsection{Implementation of Stability Components}
\label{sec:implement_architecture}

To maintain numerical values within the FP16 safety margin without sacrificing model performance, we implement Logits Soft-Capping and Sandwich Normalization.
These mechanisms cap extreme values and normalize residual branches, respectively.

\paragraph{Logits Soft-Capping.}
Standard linear layers in Large Language Models often produce logits that grow unbounded during training, causing the Softmax function to saturate and gradients to vanish or explode.
Soft-capping addresses this by squashing the logits into a fixed range using the hyperbolic tangent ($\tanh$) function before scaling them back.
Formally, given the raw logits $x$ and a capping threshold $\sigma$ (e.g., $30.0$ or $50.0$), the capped logits $x'$ are computed as:
\begin{equation}
    x' = \sigma \cdot \tanh\left(\frac{x}{\sigma}\right)
\end{equation}
In our implementation, we apply this transformation to the output logits of the language model head.
This ensures that the input to the cross-entropy loss remains within the range $(-\sigma, \sigma)$, preventing logits from exceeding the FP16 maximum value while preserving the relative order of probabilities.

\paragraph{Sandwich Normalization.}
In the standard Pre-Norm Transformer architecture, the input $x$ is normalized before the sub-layer (Attention or Feed-Forward Network), and the output is added directly to the residual stream: $x_{l+1} = x_l + F(\text{Norm}(x_l))$.
While effective, this allows the magnitude of the residual stream $x$ to grow monotonically with depth, potentially destabilizing deep networks.
Sandwich Normalization introduces an additional normalization layer explicitly on the output of the sub-layer branch before the residual addition.
The modified update rule for a block containing a sub-layer $F$ (e.g., Self-Attention or MLP) is defined as:
\begin{equation}
    x_{l+1} = x_l + \text{Norm}_{\text{post}}\left( F(\text{Norm}_{\text{pre}}(x_l)) \right)
\end{equation}
In our implementation, we apply this strictly to the residual branches.
This ensures that the contribution of each layer has unit variance, preventing the accumulation of extreme activation values as the network depth increases.

\subsection{Training Configuration}

In \Cref{tab:training_config}, we present the details of our training hyperparameter configuration in three parts:
\begin{itemize}
    \item For the model architecture, we primarily follow Qwen3-1.7B~\citep{qwen3} and adopt the vocabulary from the Qwen series~\citep{qwen2,qwen3}. We use $\theta = 10000$ for RoPE~\citep{su2024rope} to support a context length of 4K. The Soft-Capping threshold is set to 30.0, as discussed in \Cref{sec:architecture_stability}.
    \item We use AdamW as the optimizer with $\beta_1 = 0.9$ and $\beta_2 = 0.95$. We adopt a $\mu$P with base dimension of 896 and set the learning rate to $5\times10^{-3}$ for Phase~1 and $3\times10^{-3}$ thereafter before decay.
    \item To support FP16 training, we use dynamic loss scaling with a factor of 2 and a window of 20 to handle widely varying gradient scales.
\end{itemize}
This detailed configuration facilitates the reproduction of our training run.

\begin{table}[htbp]
\centering
\caption{Training Hyperparameter Configuration.}
\label{tab:training_config}
\begin{tabular}{@{}llr@{}}
\toprule
\textbf{Category} & \textbf{Parameter} & \textbf{Value} \\
\midrule
\multicolumn{3}{l}{\textbf{Model Architecture}} \\
& Sequence Length & 4096 \\
& Hidden Size & 2048 \\
& FFN Dimension & 6144 \\
& Number of Layers & 28 \\
& Number of Attention Heads & 16 \\
& Number of KV Heads (GQA) & 8 \\
& Vocabulary Size & 151936 \\
& Rotary $\theta$ & 10000.0 \\
& Logit Soft-capping threshold & 30.0 \\
& Initialization Std & 0.018 \\
\midrule
\multicolumn{3}{l}{\textbf{Optimizer Configuration}} \\
& Optimizer Type & AdamW \\
& Learning Rate (Phase 1) & $5\times 10^{-3}$ \\
& Learning Rate (Phase 2+) & $3\times 10^{-3}$ \\
& Batch Size & 2048 \\
& $\beta_1$ & 0.9 \\
& $\beta_2$ & 0.95 \\
& $\epsilon$ & 1e-8 \\
& Weight Decay & 0.1 \\
& Warmup Steps & 5000 \\
& $\mu$P Width Base & 896 \\
\midrule
\multicolumn{3}{l}{\textbf{Loss Scaling (Dynamic)}} \\
& Scale Factor & 2 \\
& Scale Window & 20 \\
& Minimum Loss Scale & 524288 \\
\bottomrule
\end{tabular}
\end{table}

\subsection{Model Average}\label{sec:model_average}

\begin{table}[htbp]
\centering
\caption{Model Performance Across Checkpoints.}
\label{tab:checkpoint_performance}
\begin{narrowtblr}{c c c c c c}
Ckpt Step & ARC-Challenge & ARC-Easy & CSQA & PIQA & Average\\
260632 & 64.41 & 82.72 & 65.93 & 73.39 & 71.61\\
261032 & 65.42 & 82.19 & 65.36 & 73.72 & 71.67\\
261432 & 63.05 & 81.48 & 65.68 & 74.59 & 71.2\\
261832 & 65.42 & 81.83 & 64.78 & 73.56 & 71.40\\
262232 & 61.36 & 83.25 & 65.77 & 73.78 & 71.04\\
262632 & 65.76 & 82.01 & 66.34 & 73.5 & 71.90\\
263032 & 63.73 & 80.78 & 66.42 & 73.78 & 71.17\\
263132 & 62.71 & 80.6 & 66.09 & 73.88 & 70.82\\
\end{narrowtblr}
\end{table}

Following recent work~\citep{luo2025learningratedecaywastes}, we average the near-end checkpoints to reduce variance and consolidate learned knowledge and capabilities.
We first evaluate the last eight checkpoints on a subset of lightweight benchmarks, as shown in \Cref{tab:checkpoint_performance}. Consecutive checkpoints are spaced 400 steps apart, corresponding to 3.36B tokens.
These checkpoints fluctuate during training and do not exhibit a clear upward or downward trend. Therefore, we apply simple model averaging~\citep{li2025modelmergingpretraininglarge}, directly averaging the last eight checkpoints to obtain the final model.

\subsection{Reference Experiments for Quantile Benchmarking}
\label{sec:referece_quantile_benchmarking}

We conduct quantile benchmarking experiments across two primary scenarios: training from scratch and continual training from checkpoints. For each experiment, given a target quantile $p\%$, we select the data partition above the $p\%$ threshold, comprising roughly 10B tokens, which are then used for the respective training scenarios.

\paragraph{Training from Scratch.}
In the training-from-scratch scenario, we train a model with the Qwen3-0.6B architecture~\citep{qwen3}. Following the default configuration in \Cref{tab:training_config}, we conduct a small-scale experiment using the settings detailed in \Cref{tab:train_from_scratch}, training over approximately 8.4B tokens from the quantile data chunks. We employ a constant learning rate schedule with a sufficiently long warmup phase to ensure stable training dynamics.

\paragraph{Continual Training.}
In the continual training scenario, we resume from a checkpoint previously trained on approximately 367B tokens of the deduplicated DCLM-Baseline dataset. The model adopts the Qwen2.5-0.5B architecture~\citep{qwen2025qwen25technicalreport}. We then train over approximately 8.4B tokens from the quantile data chunks using the configuration specified in \Cref{tab:continual_training}. For these experiments, we linearly decay the learning rate from a peak value of $1\times10^{-3}$ to a final value of $1\times10^{-5}$.

\paragraph{Consistency Across Scenarios.}
As illustrated in \Cref{fig:quantile_benchmark_understanding,fig:quantile_benchmarks_knowledge}, the benchmarking results exhibit strong alignment between the training-from-scratch and continual training experiments. This consistency persists for evaluations on both the DCLM-Baseline and Fineweb-Edu datasets, despite resuming from a checkpoint trained exclusively on the deduplicated DCLM-Baseline dataset. This observation supports the robustness of our quantile-based data selection approach across different training paradigms.

\paragraph{Compute Cost Comparison.} We use the DCLM-Baseline as a reference for comparison, as discussed in \Cref{sec:data_benchmark}. Following previous work~\cite{kaplan2025scaling}, we estimate the compute budget using the formula $C=6ND$, where $N$ represents the number of model parameters and $D$ represents the data size. Consequently, the compute cost for the 0.6B model trained on 42B tokens is approximately $(0.6\times 42) / (2\times 609)\approx 2.07\%$ relative to the baseline. Similarly, we conclude that this accounts for less than 0.6\% of the total pretraining budget.

\begin{table}[htbp]
\centering
\caption{Training Hyperparameter Configuration for Quantile Benchmarking: Training from Scratch.}
\label{tab:train_from_scratch}
\begin{tabular}{@{}lr@{}}
\toprule 
\textbf{Parameter} & \textbf{Value} \\
\midrule
Learning Rate & $1\times 10^{-3}$ \\
Batch Size & 512 \\
Warmup Steps & 400 \\
Total Steps & 4000 \\
\bottomrule
\end{tabular}
\end{table}

\begin{table}[htbp]
\centering
\caption{Training Hyperparameter Configuration for Quantile Benchmarking: Continual Training.}
\label{tab:continual_training}
\begin{tabular}{@{}lr@{}}
\toprule 
\textbf{Parameter} & \textbf{Value} \\
\midrule
Peak Learning Rate & $1\times 10^{-3}$ \\
Final Learning Rate & $1\times 10^{-5}$ \\
Batch Size & 2048 \\
Total Steps & 1000 \\
\bottomrule
\end{tabular}
\end{table}

\subsection{Reference Experiments for Repetition and Curriculum Model Averaging}\label{sec:reference_repetition_curriculum}

These experiments primarily follow the experimental framework established in CMA~\citep{luo2025learningratedecaywastes}. We use a model with the Qwen2.5-1.5B architecture without tied embeddings and train on a subset of the first shard of the DCLM-Baseline dataset.

\paragraph{Baseline Configuration.}
The baseline experiment adopts uniform data ordering and employs a Warmup-Stable-Decay (WSD) learning rate schedule with a $1$-sqrt decay function~\citep{hagele,tian2025wsmdecayfreelearningrate}, decaying to a near-zero final learning rate. The detailed experimental configuration is provided in \Cref{tab:1b5_setting}.

\paragraph{High-Quality Data Utilization Strategies.}
To investigate effective high-quality data utilization, we explore two complementary approaches:
\begin{itemize}
    \item[(1)] \textbf{Repetition Strategy:} We repeat high-quality data partitions for various top-$k$ retention ratios, matching the computational FLOPs of the single-pass baseline experiment for fair performance comparison.
    \item[(2)] \textbf{Curriculum with Model Averaging:} We adopt CMA/CDMA\footnote{We do not distinguish these variants in our context, and refer to both as \textit{CMA}. By definition, the CMA method in \Cref{tab:curriculum_comparison} corresponds to the CDMA variant, which retains LR decay.}~\citep{luo2025learningratedecaywastes}, which integrates curriculum learning with either no or moderate LR decay, accompanied by model averaging over the final checkpoints.
\end{itemize}

\begin{table}[htbp]
\centering
\caption{Training Hyperparameter Configuration for Baseline and Repetition.}
\label{tab:1b5_setting}
\begin{tabular}{@{}lr@{}}
\toprule 
\textbf{Parameter} & \textbf{Value} \\
\midrule
Peak Learning Rate & $3\times 10^{-3}$ \\
Final Learning Rate & $1\times 10^{-5}$ \\
Batch Size & 512 \\
Total Steps & 15,375 \\
Decay Steps & 2,875 \\
Warmup Steps & 768 \\
\bottomrule
\end{tabular}
\end{table}

\begin{table}[htbp]
\centering
\caption{Training Hyperparameter Configuration for Curriculum Model Average.}
\label{tab:1b5_curriculum_setting}
\begin{tabular}{@{}lr@{}}
\toprule 
\textbf{Parameter} & \textbf{Value} \\
\midrule
Peak Learning Rate & $3\times 10^{-3}$ \\
Final Learning Rate & $1\times 10^{-3}$ \\
Batch Size & 512 \\
Total Steps & 15,375 \\
Decay Steps & 2,875 \\
Warmup Steps & 768 \\
Checkpoint Number & 6\\
Decay Factor of EMA ($\alpha$) & 0.2\\
Checkpoint Interval & 0.21B\\
\bottomrule
\end{tabular}
\end{table}

\paragraph{Experimental Variants.}
The repetition experiments follow identical settings to the baseline, differing only in dataset construction. For the curriculum experiments, we use a higher final learning rate of $1\times10^{-3}$ and perform an exponential moving average (EMA) over the final six checkpoints (the last-step checkpoint is weighted by $(1-\alpha)$ relative to the current-step checkpoint, where $\alpha$ is the decay factor), replicating the methodology from CMA~\citep{luo2025learningratedecaywastes}.

\paragraph{Evaluation Settings.}
In \Cref{tab:curriculum_comparison}, we evaluate performance on a high–signal-to–noise-ratio benchmark subset (\textit{Core} in \Cref{tab:curriculum_comparison}) comprising MMLU~\citep{mmlu}, ARC~\citep{arc}, and CSQA~\citep{commonsenseqa}, following established practices in prior work~\citep{heineman2025signalnoiseframeworkreducing,luo2025learningratedecaywastes}. These benchmarks provide strong discriminative power for identifying performance differences between training approaches.

\section[Model Perf. across Benchmarks (Full)]{Model Performance across Benchmarks (Full Table)}
\label{sec:model_perf_full}

\Cref{tab:model_comparison_full} merges the results from both \Cref{tab:lang_math_code,tab:reasoning_knowledge}, providing a complete evaluation of the models across all target capability dimensions. Because the models differ in total parameters and non-embedding parameters, we present performance–parameter visualizations in \Cref{fig:model_perf_comp,fig:performance_comparison_on_nonembedding}. These plots show that \sys lies on the frontier of fully open-source models.

\begin{landscape}
\begin{table}[p]
\centering
\caption{Comparison of Model Performance across Various Benchmarks.}
\label{tab:model_comparison_full}
\resizebox{\linewidth}{!}{%

\begin{tabular}{lccccccccccccccccc}
\toprule
\multirow{2}{*}{\textbf{Model Name}} & \multirow{2}{*}{\textbf{Params}} & \multicolumn{2}{c}{\textbf{Math}} & \multicolumn{2}{c}{\textbf{Code}} & \multicolumn{2}{c}{\textbf{Chinese}} & \multicolumn{9}{c}{\textbf{Reasoning \& Knowledge}} & \multirow{2}{*}{\textbf{Avg.}} \\
\cmidrule(lr){3-4} \cmidrule(lr){5-6} \cmidrule(lr){7-8} \cmidrule(lr){9-17}
 &  & GSM8K & MATH & sanitized\_MBPP & HumanEval & C-Eval & CMMLU & MMLU & ARC-C & ARC-E & BoolQ & CSQA & HSwag & PIQA & SocIQ & Wino & \\
\midrule

\multicolumn{18}{l}{\textit{\textbf{Open-Weight SOTA Models}}} \\
\rowcolor{openweight} Qwen2-1.5B & 1.5B & 58.50 & 21.70 & 50.58 & 31.10 & 71.29 & 70.62 & 56.36 & 70.17 & 83.60 & 71.90 & 70.52 & 60.77 & 75.73 & 63.46 & 59.83 & 61.08 \\
\rowcolor{openweight} Qwen2.5-1.5B & 1.5B & 68.50 & 35.00 & 58.37 & 37.20 & 68.63 & 68.01 & 61.56 & 79.32 & 90.48 & 76.39 & 75.10 & 64.18 & 76.17 & 64.94 & 59.67 & 65.57 \\
\rowcolor{openweight} Qwen2.5-3B   & 3B   & 79.10 & 42.60 & 66.54 & 42.10 & 74.65 & 73.92 & 66.86 & 86.44 & 92.59 & 83.88 & 76.09 & 73.85 & 81.45 & 69.40 & 63.69 & 71.54 \\
\rowcolor{openweight} Qwen3-0.6B   & 0.6B & 59.59 & 32.44 & 51.75 & 29.88 & 57.03 & 52.36 & 55.09 & 68.14 & 84.48 & 69.05 & 61.18 & 48.51 & 69.97 & 61.51 & 55.64 & 57.11 \\
\rowcolor{openweight} Qwen3-1.7B   & 1.7B & 75.44 & 43.50 & 64.20 & 52.44 & 66.70 & 66.55 & 65.35 & 80.34 & 91.89 & 79.82 & 74.61 & 60.76 & 77.20 & 68.58 & 59.27 & 68.44 \\
\rowcolor{openweight} Qwen3-4B     & 4B   & 87.79 & 54.1 & 74.32 & 62.2 & 78.5 & 77.01 & 75.78 & 89.83 & 97.53 & 86.09 & 81.9 & 79.46 & 84.98 & 75.59 & 65.43 & 78.03 \\
\rowcolor{openweight} gemma2-2B    & 2B   & 23.90 & 15.00 & 38.91 & 17.70 & 41.35 & 39.63 & 55.20 & 66.44 & 82.54 & 72.42 & 69.45 & 66.20 & 78.89 & 65.92 & 65.35 & 53.26 \\
\rowcolor{openweight} llama-3.2-1B & 1B   &  44.40 & 30.60 & 34.63 & 18.90 & 29.82 & 31.03 & 37.74 & 36.95 & 70.55 & 67.43 & 62.82 & 60.20 & 74.92 & 50.61 & 58.17 & 47.25 \\
\rowcolor{openweight} llama-3.2-3B & 3B   & 77.70 & 48.00 & 49.42 & 29.88 & 45.67 & 44.33 & 57.87 & 72.20 & 83.95 & 76.73 & 70.35 & 71.06 & 79.05 & 64.33 & 64.09 & 62.31 \\

\midrule

\multicolumn{18}{l}{\textit{\textbf{Fully-Open SOTA Models}}} \\
\rowcolor{fullyopen} SmolLM2-1.7B       & 1.7B & 31.10 & 11.60 & 49.42 & 22.60 & 35.06 & 34.03 & 51.99 & 59.66 & 82.72 & 69.85 & 67.16 & 65.30 & 78.51 & 60.18 & 59.12 & 51.89 \\
\rowcolor{fullyopen} OLMo-2-0425-1B     & 1B   & 68.30 & 20.70 & 15.56 & 6.71 & 30.53 & 28.62 & 44.25 & 47.46 & 76.72 & 70.55 & 65.60 & 61.61 & 76.44 & 55.53 & 60.38 & 48.60 \\
\rowcolor{fullyopen} YuLan-Mini-2.4B    & 2.4B & 66.65 & 27.12 & 62.26 & 61.60 & 52.32 & 48.14 & 51.76 & 64.75 & 82.54 & 78.59 & 66.18 & 61.20 & 77.31 & 63.25 & 61.88 & 61.70 \\
\rowcolor{fullyopen} SmolLM3-3B         & 3B   & 67.63 & 46.10 & 62.26 & 39.63 & 50.84 & 49.35 & 63.04 & 77.29 & 88.54 & 76.12 & 70.52 & 69.20 & 79.05 & 65.25 & 64.40 & 64.61 \\

\midrule

\multicolumn{18}{l}{\textit{\textbf{Ours}}} \\
\rowcolor{ours} PCMind-2.1-Kaiyuan-2B & 2B & 51.33 & 30.34 & 56.42 & 42.68 & 46.30 & 49.25 & 53.90 & 66.10 & 82.89 & 78.53 & 67.40 & 58.13 & 74.37 & 62.59 & 65.75 & 59.07 \\

\bottomrule
\end{tabular}%

}
\end{table}
\end{landscape}

\end{document}